\documentclass[12pt]{article}
\usepackage[a4paper,margin=1in]{geometry}
\usepackage{setspace}
\usepackage{amsmath, amssymb}
\usepackage{float}
\usepackage{titlesec}
\usepackage{hyperref}
\usepackage{graphicx}
\usepackage{geometry}
\usepackage{caption}
\usepackage{fancyhdr}
\usepackage{algorithm}
\usepackage{algorithmicx}
\usepackage{algpseudocode}
\usepackage{enumitem}
\usepackage{booktabs}
\usepackage{caption}
\usepackage[table]{xcolor}
\usepackage{colortbl}
\usepackage[utf8]{inputenc}
\usepackage{subcaption}
\usepackage{pdflscape}
\usepackage{float}
\usepackage[margin=1in]{geometry}
\usepackage[toc,page]{appendix} 

\titleformat{\section}{\large\bfseries\uppercase}{\thesection.}{1em}{}
\titleformat{\subsection}{\normalsize\bfseries}{\thesubsection.}{1em}{}

\begin{document}

\begin{center}
    {\Large \textbf{Solving N-Queen Problem using Las Vegas Algorithm with State pruning}}\\[10pt]
    {\large Susmita Sharma \textsuperscript{*}, Aayush Shrestha, Prashant Timalsina, Sitasma Thapa, Prakash Poudyal }\\[5pt]
    {\normalsize \textsuperscript{*,1,2} Department of Mathematics, Department of Computer Science and Engineering, Kathmandu University, Dhulikhel, Kavre \\
    Corresponding Author: Prakash Poudyal	
    Email: \texttt{prakash@ku.edu.np}}
\end{center}

\vspace{0.5cm}
\section*{Abstract}
\small{ 
The N-Queens problem, placing all $N$ queens in a $N$ x$N$ chessboard where none attack the other, is a classic problem for constraint satisfaction algorithms. While complete methods like backtracking guarantee a solution, their exponential time complexity makes them impractical for large-scale instances thus,  stochastic approaches, such as Las Vegas algorithm, are preferred. While it offers faster approximate solutions, it suffers from significant performance variance due to random placement of queens on the board. This research introduces a hybrid algorithm built on top of the standard Las Vegas framework through iterative pruning, dynamically eliminating invalid placements during the random assignment phase, thus this  method effectively reduces the search space. The analysis results that traditional backtracking scales poorly with increasing N. In contrast, the proposed technique consistently generates valid solutions more rapidly, establishing it as a superior alternative to use  where a single, timely solution is preferred over completeness. Although large N causes some performance variability, the algorithm demonstrates a highly effective trade-off between computational cost and solution fidelity, making it particularly suited for resource-constrained computing environments.}
\\
\\
\noindent \textbf{Keywords:} N-Queens Problem, Las Vegas Algorithm, Stochastic Solutions, Backtracking, Pruning 

\vspace{7mm}

\newpage
\section{Introduction}

The N-Queens problem represents one of the most enduring challenges in combinatorial optimization and constraint satisfaction problems (CSPs), with applications spanning artificial intelligence, scheduling, and resource allocation. This classical problem requires placing $N$ chess queens on an $N \times N$ chessboard such that no two queens threaten each other horizontally, vertically, or diagonally. While the problem appears deceptively simple, its computational complexity has made it a benchmark for evaluating algorithmic approaches across multiple domains. The $n$-Queens problem is solved by finding a way to place $n$ queens on an $n \times n$ chessboard so that none attack each other—is a long-studied combinatorial and constraint satisfaction challenge (Bell \& Stevens, 2009 \cite{Bell2009}). Although deterministic constructions exist (Hoffman et al., 1969)\cite{Hoffman1969}, enumeration or solution for arbitrary $n$ becomes intractable as complexity grows exponentially (Bell \& Stevens, 2009 \cite{Bell2009}). Advanced constraint techniques and heuristics help manage this exponential growth through state pruning mechanisms that eliminate attacked squares from future consideration, substantially reducing the search tree.\\

Traditional approaches have predominantly relied on backtracking algorithms, which systematically explore the solution space by incrementally building partial solutions and abandoning paths that cannot lead to valid configurations. However, as the problem size increases, the exponential growth of the state space—ranging from $2^{N^2}$ for unrestricted placements to $N!$ for column-wise restrictions—creates significant computational challenges. This exponential complexity necessitates the development of more efficient algorithmic strategies that can handle larger problem instances within reasonable time constraints.\\

Recent advances in randomized algorithms have demonstrated significant promise in addressing computationally intractable problems. Among these, Las Vegas algorithms offer particularly attractive properties for constraint satisfaction problems, guaranteeing correct solutions while accepting variable execution times. Unlike Monte Carlo algorithms that trade correctness for bounded execution time, Las Vegas algorithms ensure solution accuracy while leveraging randomization to achieve expected polynomial-time performance. State space reduction techniques have emerged as complementary approaches to traditional search methods, focusing on minimizing the search space size through intelligent pruning and constraint propagation. These techniques eliminate infeasible regions of the solution space before exhaustive exploration, potentially reducing computational complexity from exponential to polynomial time for certain problem classes. A key concept in efficient search is state pruning or forward-checking: once a queen is placed, all attacked squares (row, column, diagonals) are eliminated from future consideration, reducing the search tree substantially.\\

This research investigates the integration of state space reduction techniques with Las Vegas algorithms for solving the N-Queens problem, comparing this hybrid approach against traditional backtracking methods. The significance of this work lies in addressing the fundamental trade-off between solution guarantees and computational efficiency that has characterized N-Queens research for over a century.\\

The paper presents a combination of Las Vegas algorithm approach with state space reduction techniques to tackle the N-Queens problem more efficiently. The literature review highlights past approaches, from classical backtracking to probabilistic and current heuristic methods, and their limitations. Following the literature review, the methodology section presents the working and analysis along with an example of our modified Las Vegas algorithm with pruning.\\

\section{Literature Review}

The N-Queens problem was first formalized by Max Bezzel in 1848, with Franz Nauck providing the initial solutions in 1850. The problem's mathematical properties have been extensively studied, with research establishing that solutions exist for all natural numbers $N$ except $N = 2$ and $N = 3$. The asymptotic growth rate of solutions follows approximately $(0.143N)^N$, indicating the exponential nature of the solution space.\\

Early explorations in N-Queens solving leaned heavily on deterministic backtracking approaches, which guarantee completeness but often become computationally prohibitive in vast search spaces. Arteaga et al. (2022) \cite{Arteaga2022} conducted comprehensive comparisons of brute-force, branch-and-bound, and integer programming approaches up to $N=17$, demonstrating how computational costs escalate dramatically with problem size. To address these limitations, state pruning—also known as forward checking—emerged as a crucial optimization technique. Haralick and Elliott (1980) \cite{Haralick1980} pioneered the integration of constraint propagation with backtracking search, where placing a queen immediately marks off attacked squares, dramatically reducing subsequent branching factors. Russell and Norvig (2020) \cite{Russell2020} further highlighted how pairing forward checking with heuristics such as minimum remaining values (MRV) can reduce backtracking operations by orders of magnitude, transforming previously infeasible tasks into manageable computational problems. Deterministic construction methods have also contributed significantly to the field. Hoffman et al. (1969) \cite{Hoffman1969}  developed an $O(n)$-time construction method applicable to most values of $n$ greater than 3, while Bernhardsson (1991) \cite{Bernhardsson1991} extended explicit placement algorithms to cover all valid $n$ values. However, these deterministic approaches, while efficient for finding single solutions, do not facilitate complete enumeration or provide insights into the full solution space structure.\\

The limitations of deterministic methods for large-scale instances led researchers to explore probabilistic approaches.  Sosic and Gu (1990) \cite{Sosic1990} introduced groundbreaking polynomial-time algorithms employing random permutations of queens with iterative improvement strategies, capable of handling instances up to $N=500,000$. Their approach utilized efficient diagonal tracking mechanisms to identify and resolve conflicts in constant time, demonstrating the practical scalability of randomized methods. Building upon this foundation, Sosic and Gu (1991) \cite{Sosic1991} developed enhanced probabilistic algorithms, including QS1's row-by-row random placement strategy with restart mechanisms, achieving near-linear performance for million-queen problems through forward checking techniques. These methods established important precedents for Las Vegas algorithmic approaches to constraint satisfaction problems. Rolfe (2006) \cite{Rolfe2006} later adapted these concepts for educational contexts, developing two Las Vegas variants: permutation-based swap algorithms and GridSearch's random safe placement per row with restart capabilities. These simplified approaches maintained effectiveness for instances up to $N=1000$ while providing pedagogically accessible implementations.\\

Local search heuristics have demonstrated remarkable success in solving large-scale N-Queens instances. Minton et al. (1992) \cite{Minton1992} pioneered the min-conflicts approach for constraint satisfaction problems, where queens are iteratively relocated to positions that minimize constraint violations. This method employs randomization to escape local minima and has proven capable of solving million-queen boards within minutes.  \cite{Sosic1994} extended conflict-minimization search techniques, achieving solutions for 1000-queen problems in seconds. These approaches, while not strictly Las Vegas algorithms due to potential stalling behavior requiring restarts, effectively demonstrate the power of randomization in pruning unproductive search paths. Metaheuristic approaches have introduced nature-inspired optimization techniques to the N-Queens domain. Al-Gburi et al. (2018) \cite{AlGburi2018} developed hybrid bat-genetic algorithms that outperformed individual genetic algorithm implementations on $500 \times 500$ boards, leveraging echolocation-inspired search mechanisms combined with evolutionary crossover operations to reduce iteration requirements and improve solution reliability. Erbaş et al. (1992) \cite{Erbas1992} explored probabilistic beam search perspectives, employing narrowed beam widths to prune weak search paths while solving instances up to $N=500$. These methods demonstrate the continued evolution of metaheuristic approaches in addressing large-scale combinatorial optimization problems.\\

Modern constraint programming techniques have significantly advanced N-Queens solving capabilities.Van Beek (2006) \cite{vanBeek2006} surveyed the evolution of backtracking algorithms, demonstrating how randomized variable ordering and restart strategies transform traditional backtracking into Las Vegas algorithms, with arc consistency techniques pruning inconsistencies before search initiation. Gomes et al. (2008) \cite{Gomes2008}established connections between N-Queens solving and satisfiability (SAT) solvers, where clause learning and geometric restart strategies enable unit propagation-based pruning for large instances. These advanced constraint programming methods represent the current state-of-the-art for exact solution approaches. Recent theoretical developments have focused on solution counting and bound refinement. Polson and Sokolov (2024) \cite{Polson2024} addressed counting challenges using Monte Carlo methods to achieve polynomial-time estimates through quantile reordering techniques. Luria and Simkin (2021)\cite{Luria2021} contributed tighter lower bounds on solution numbers, refining asymptotic understanding of the solution space structure.\\

Comparative studies reveal distinct performance characteristics across different algorithmic approaches. Las Vegas algorithms with pruning generally outperform pure backtracking methods by avoiding exhaustive search patterns, though restart frequency varies significantly based on problem structure. Constraint programming methods can efficiently solve instances up to $N=2000$ \cite{vanBeek2006}, but Las Vegas approaches maintain advantages in implementation simplicity and educational accessibility.
Min-conflicts and related local search methods often achieve superior linear scaling properties (Minton et al., 1992; Sosic \& Gu, 1994)\cite{Minton1992, Sosic1994}, while metaheuristic approaches typically exhibit slower convergence but demonstrate greater adaptability (Al-Gburi et al., 2018)\cite{AlGburi2018}. Arteaga et al. (2022) \cite{Arteaga2022} benchmark studies for smaller instances suggest potential advantages for Las Vegas approaches in single-solution finding scenarios.
The evolution from Sosic and Gu's (1990) \cite{Sosic1990} pioneering breakthroughs to contemporary hybrid approaches illustrates how randomization and pruning techniques continue to address the combinatorial complexity inherent in the N-Queens problem, maintaining relevance both as a computational challenge and as a pedagogical tool for algorithm design education.

\section{Algorithm}
The algorithm is divided into two function, the first is used to calculate and prune targeted position after placing a queen and the other is the main working of Las Vegas algorithm which facilitates random placement of queens.

\subsection{Invalid Points Algorithm}

The Invalid Points Algorithm, \ref{alg:invalid_points} is used to identify all positions on an $n \text{ x } n$ chessboard that become unsafe when a queen is placed on the board. Once a queen occupies a square at position $(i, j)$, it threatens all other squares in the same row, the same column, and along both diagonals passing through that point. To determine these attacked positions, the algorithm first initializes an empty set. It then iterates through all indices from $0$ to $n-1$, adding to the set every position that lies in the same row , the same column, the negative diagonal, and the positive diagonal. After finding all the unsafe positions, the algorithm removes the queen’s own position from the set.\\

Finally, the algorithm filters out any positions that fall outside the boundaries of the board, ensuring that both row and column indices lie within the range $0$ to $n-1$. Then the resulting set of invalid positions is returned and used to restrict future queen placements, effectively reducing the size of the search space. Since the algorithm performs checks across rows, columns, and diagonals in a single pass, its overall time complexity is $O(n^2)$..

\begin{algorithm}[H]
\caption{Invalid Points Calculation}
\label{alg:invalid_points}
\begin{algorithmic}[1]
    \Require $n$, $i$, $j$ (Board size and queen position)
    \Ensure Set of invalid positions for a queen placed at $(i, j)$
    
    \State Initialize $\textit{points} \gets \emptyset$
    \For{$x = 0$ to $n-1$}
        \State Add $(x, j)$ to $\textit{points}$  \Comment{Mark all positions in column $j$}
        \State Add $(i, x)$ to $\textit{points}$  \Comment{Mark all positions in row $i$}
        \State Add $(x, i + j - x)$ to $\textit{points}$  \Comment{Mark negative diagonal}
        \State Add $(x, x - i + j)$ to $\textit{points}$  \Comment{Mark positive diagonal}
    \EndFor
    \State Remove $(i, j)$ from $\textit{points}$  \Comment{Remove the queen's position}
    \State Filter $\textit{points}$ to retain only $(p,q)$ where $0 \leq p, q < n$
    \Return $\textit{points}$
\end{algorithmic}
\end{algorithm}

\subsection{Function: Las Vegas Algorithm}

The Las Vegas Algorithm, \ref{alg:las_vegas} solves the N-Queens problem by randomly placing queens on an $n \text{ x }n$ chessboard while making sure that no two queens can attack each other. 
\begin{algorithm}[H]
\caption{Las Vegas Algorithm for $n$-Queens Problem}
\label{alg:las_vegas}
\begin{algorithmic}[1]
    \Require $n$ (Size of the board)
    \Ensure An $n \times n$ matrix with queens placed
    
    \State Define function \texttt{create\_boolean\_array}($n$)
    \State \quad Return an $n \times n$ matrix of ones
    
    \State Define function \texttt{las\_vegas}($n$)
    \State \quad $a \gets \text{True}$
    \While{$a$ is True}
        \State \quad Initialize $\textit{result} \gets$ create a boolean array of size $n \times n$
        \State \quad Initialize $\textit{valid\_space} \gets$ list of all indices in $n \times n$
        \For{$i = 0$ to $n-1$}
            \State \quad Select random $q\_pos$ from $\textit{valid\_space}$ and remove it
            \State \quad Compute $x = \lfloor q\_pos / n \rfloor$, $y = q\_pos \mod n$
            \State \quad Compute attack positions using \texttt{invalid\_points}($n, x, y$)
            \For{each $(p, q)$ in attack positions}
                \State \quad Set $\textit{result}[p, q] \gets 0$
                \State \quad Remove corresponding index from $\textit{valid\_space}$
            \EndFor
            \If{$\textit{valid\_space}$ is empty}
                \If{$i > n-2$}
                    \State \quad $a \gets$ False
                \EndIf
                \State \quad \textbf{break}
            \EndIf
        \EndFor
    \EndWhile
    \Return $\textit{result}$
\end{algorithmic}
\end{algorithm}
It begins with a board where all positions are marked as valid, and a one dimensional list is created containing every possible position on the board. Then, the algorithm randomly picks one valid position from this list and places a queen there. After placing a queen on the board, it uses the Invalid Points Algorithm, \ref{alg:invalid_points} to find all positions that become unsafe because of that queen. These unsafe positions are marked on the board and also removed from the list of valid positions.\\

If the algorithm runs out of valid positions before placing all n queens, it starts over and tries again. This process continues until queens are successfully placed in all rows without conflicts. 
Once a valid arrangement is found, the algorithm returns the completed board along with the number of attempts it took. Since the solution depends on how many retries are needed, the overall time complexity is $O(p \cdot n^2)$, where $p$  is the number of attempts before a solution is found.

Now we present the working example of the algorithm for the case of N=4

First we intialize a board with all position as valid,(V)

\bigskip

\begin{table}[H]
\centering

\begin{tabular}{|>{\centering\arraybackslash}p{0.7cm}
                |>{\centering\arraybackslash}p{0.7cm}
                |>{\centering\arraybackslash}p{0.7cm}
                |>{\centering\arraybackslash}p{0.7cm}|}
\hline
\cellcolor{blue!30}V &  \cellcolor{blue!30}V    &  \cellcolor{blue!30}V   &  \cellcolor{blue!30}V   \\
\hline
\cellcolor{blue!30}V   & \cellcolor{blue!30}V & \cellcolor{blue!30}V & \cellcolor{blue!30}V \\
\hline
\cellcolor{blue!30}V   & \cellcolor{blue!30}V & \cellcolor{blue!30}V & \cellcolor{blue!30}V \\
\hline
\cellcolor{blue!30}V   & \cellcolor{blue!30}V & \cellcolor{blue!30}V & \cellcolor{blue!30}V \\
\hline
\end{tabular}
\caption{Initialize a board with all position as valid.}
\label{tab:Initial_Board}
\end{table}

Next, our algorithm random selects from one of the valid position, to place the queen, Q. And the valid point algorithm \ref{alg:invalid_points} is used to invalidate the attacked positions, X.
\begin{table}[H]

\begin{tabular}{|>{\centering\arraybackslash}p{0.7cm}
                |>{\centering\arraybackslash}p{0.7cm}
                |>{\centering\arraybackslash}p{0.7cm}
                |>{\centering\arraybackslash}p{0.7cm}|}
\hline
\cellcolor{red!30}X &  \cellcolor{red!30}X    &  \cellcolor{blue!30}V   &  \cellcolor{blue!30}V   \\
\hline
\cellcolor{green!30}Q   & \cellcolor{red!30}X & \cellcolor{red!30}X & \cellcolor{red!30}X \\
\hline
\cellcolor{red!30}X   & \cellcolor{red!30}X & \cellcolor{blue!30}V & \cellcolor{blue!30}V \\
\hline
\cellcolor{red!30}X   & \cellcolor{blue!30}V & \cellcolor{red!30}X & \cellcolor{blue!30}V \\
\hline
\end{tabular}
\label{tab:Placing Queen}
\centering
\caption{Place Queen at position (1,0). Mark attacked squares in red (X).}
\end{table}

\bigskip
 The above Step 1 is then repeated by choosing another random positions that is still valid. The algorithm continues placing queens till there are at least as much valid positions as the queens to be placed. 
 This can result in two scenerios.
\begin{table}[H]

\begin{tabular}{|>{\centering\arraybackslash}p{0.7cm}
                |>{\centering\arraybackslash}p{0.7cm}
                |>{\centering\arraybackslash}p{0.7cm}
                |>{\centering\arraybackslash}p{0.7cm}|}
\hline
\cellcolor{red!30}X &  \cellcolor{red!30}X    &  \cellcolor{green!30}Q   &  \cellcolor{red!30}X   \\
\hline
\cellcolor{green!30}Q   & \cellcolor{red!30}X & \cellcolor{red!30}X & \cellcolor{red!30}X \\
\hline
\cellcolor{red!30}X   & \cellcolor{red!30}X & \cellcolor{red!30}X & \cellcolor{green!30}Q \\
\hline
\cellcolor{red!30}X   & \cellcolor{green!30}Q & \cellcolor{red!30}X & \cellcolor{red!30}X \\
\hline
\end{tabular}
\label{tab:Solved}
\centering
\caption{N queens placed on a $n$x$n$ chessboard that follows constraint.}
\end{table}

\bigskip

\begin{table}[H]
\centering

\begin{tabular}{|>{\centering\arraybackslash}p{0.7cm}
                |>{\centering\arraybackslash}p{0.7cm}
                |>{\centering\arraybackslash}p{0.7cm}
                |>{\centering\arraybackslash}p{0.7cm}|}
\hline
\cellcolor{red!30}X &  \cellcolor{red!30}X  &  \cellcolor{red!30}X   &  \cellcolor{green!30}Q      \\
\hline
\cellcolor{green!30}Q   & \cellcolor{red!30}X & \cellcolor{red!30}X & \cellcolor{red!30}X \\
\hline
\cellcolor{red!30}X   & \cellcolor{red!30}X & \cellcolor{red!30}X & \cellcolor{red!30}X \\
\hline
\cellcolor{red!30}X   & \cellcolor{green!30}Q & \cellcolor{red!30}X & \cellcolor{red!30}X \\
\hline
\end{tabular}
\caption{Less than N queens placed on a $n$x$n$ chessboard that follows constraint.}
\label{tab:dead_end}
\end{table}

\bigskip
Here on Scenario 2 in Table~\ref{tab:dead_end}, the algorithm has exhausted the valid space without solving the nqueen problem, as such it restarts the whole search from Step 1 as in Table~\ref{tab:Initial_Board}.

\subsection{Analysis}
In this section, we present a formal analysis of the above algorithm to demonstrate its correctness. The analysis uses the concept of Loop Invariant Statement which is a condition that holds true before and after each iteration of a loop. By establishing such an condition, we  show that the algorithm progresses towards its intended goal without violating essential correctness criteria \\

\textbf{Loop Invariant Statement}\\
At the start of each iteration of the loop in the las vegas function, the result matrix correctly represents the state of the board, where queens have been placed in valid positions, and all invalid positions (due to the placed queens) have been marked accordingly. Additionally, the valid space list contains only those positions that are still available for placing a new queen.\\

\textbf{Initialization}\\
The result matrix is initially filled with 1s, indicating that all positions are valid for placing a queen, and valid space list consists of index $0$ to $n^2-1$ Since before the loop begins, no queens have been placed, every spot on the board is considered valid, and the valid space list includes all available positions.\\

\textbf{Maintenance:}\\
A random position is chosen from the valid space list and then is  removed from that list. Positions that become invalid due to previously placed queens are marked as 0.  From the remaining indices in the valid space list, the loop selects the random position. If the queen needs to be placed and there is no valid solution then the loop restarts. If all queens are successfully placed in valid positions, the loop will stop; otherwise, it restarts until a valid solution is found. The result matrix will then show the valid positions where the queen is placed, with invalid positions marked accordingly. The valid space list will contain only the positions where queens can still be placed.\\

\textbf{Termination:}\\
The loop ends when all queens are successfully placed in valid positions. If no valid position is found, the loop continues as usual until a valid solution is found. As a result, the result matrix shows where each queen is placed correctly, and the valid space list includes positions still available for placing additional queens.

\section{Results and Discussion}
The performance of the proposed algorithm for the n-Queens problem was evaluated against the classical backtracking approach. Analysis was done on the basis of number of queens placed before reaching the correct configuration and each algorithm was run for each $n$ till first stable placement was found. As the Las Vegas Algorithm, \ref{alg:las_vegas} is probabilistic in nature, the algorithm was run for more than 1000 times for each $n$. Metrics such as mean, median, mode, skewness, and kurtosis of the number of those attempts were computed for the probabilistic method, while the number of attempts for backtracking was directly measured.\\ 

Histograms with kernel density estimates (KDE) for $n$ values, as in \ref{fig:las_vegas_distributions}, reveal that the probabilistic algorithm exhibits positively skewed distributions with long right tails. This indicates that while most runs converge in relatively few attempts, there exists a non-negligible probability of encountering runs requiring orders of magnitude more attempts.\\

The skewness values consistently lie around 1.9–2.1, with kurtosis between 5.1–6.6, suggesting heavy-tailed, leptokurtic distributions. This implies a concentration of outcomes around the mean but with frequent outliers on the higher end. The best-fitting distributions for various $n$ include gamma, Weibull (minimum), and Pareto distributions, indicating that the algorithm’s runtime is probabilistically bounded but not safe from to rare, extreme delays.

\begin{figure}[H]
\centering
\begin{subfigure}[b]{0.3\textwidth}
    \includegraphics[width=\textwidth]{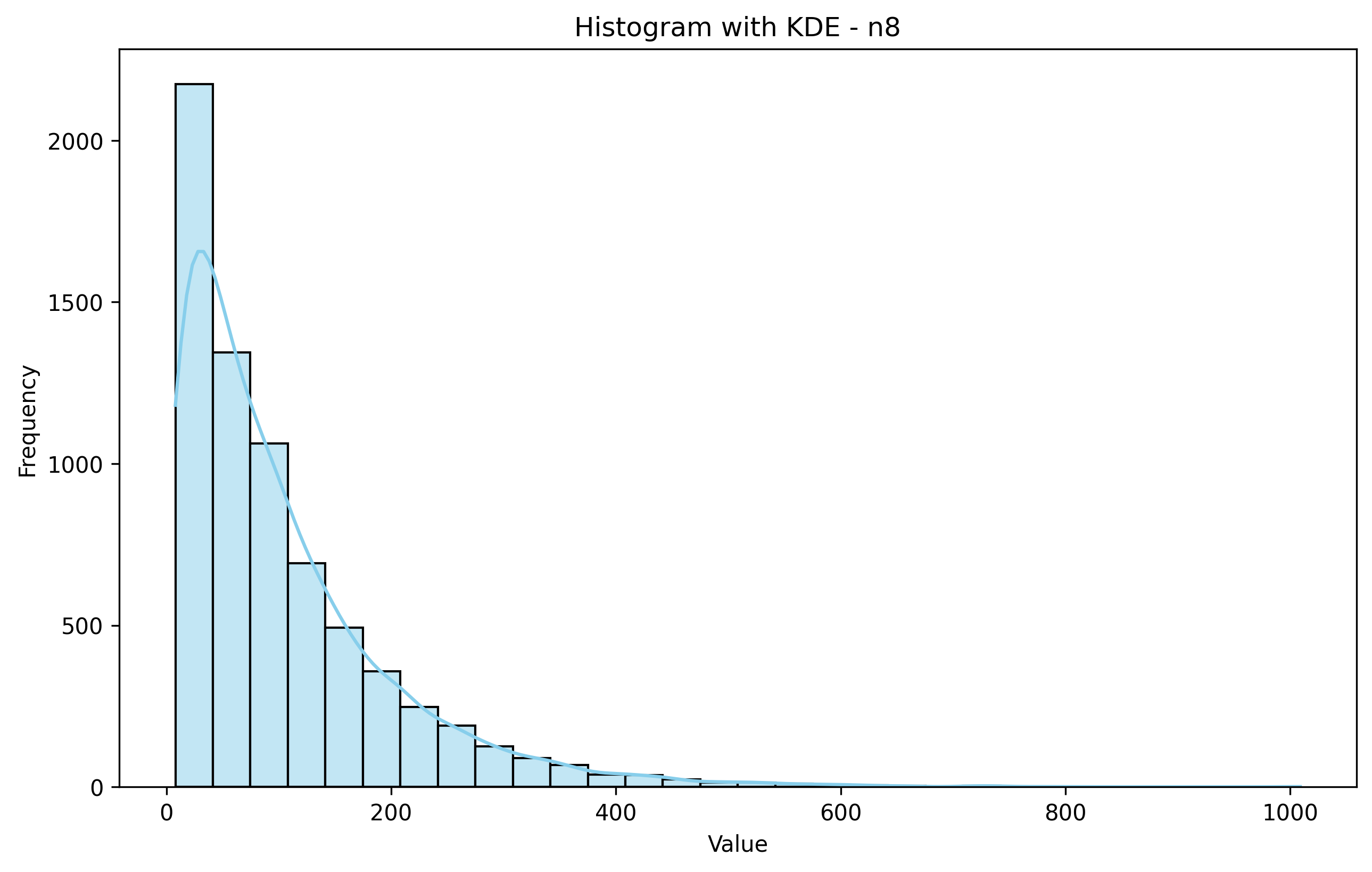}
    \caption{Las Vegas Attempt distribution for n = 8}
\end{subfigure}
\hfill
\begin{subfigure}[b]{0.3\textwidth}
    \includegraphics[width=\textwidth]{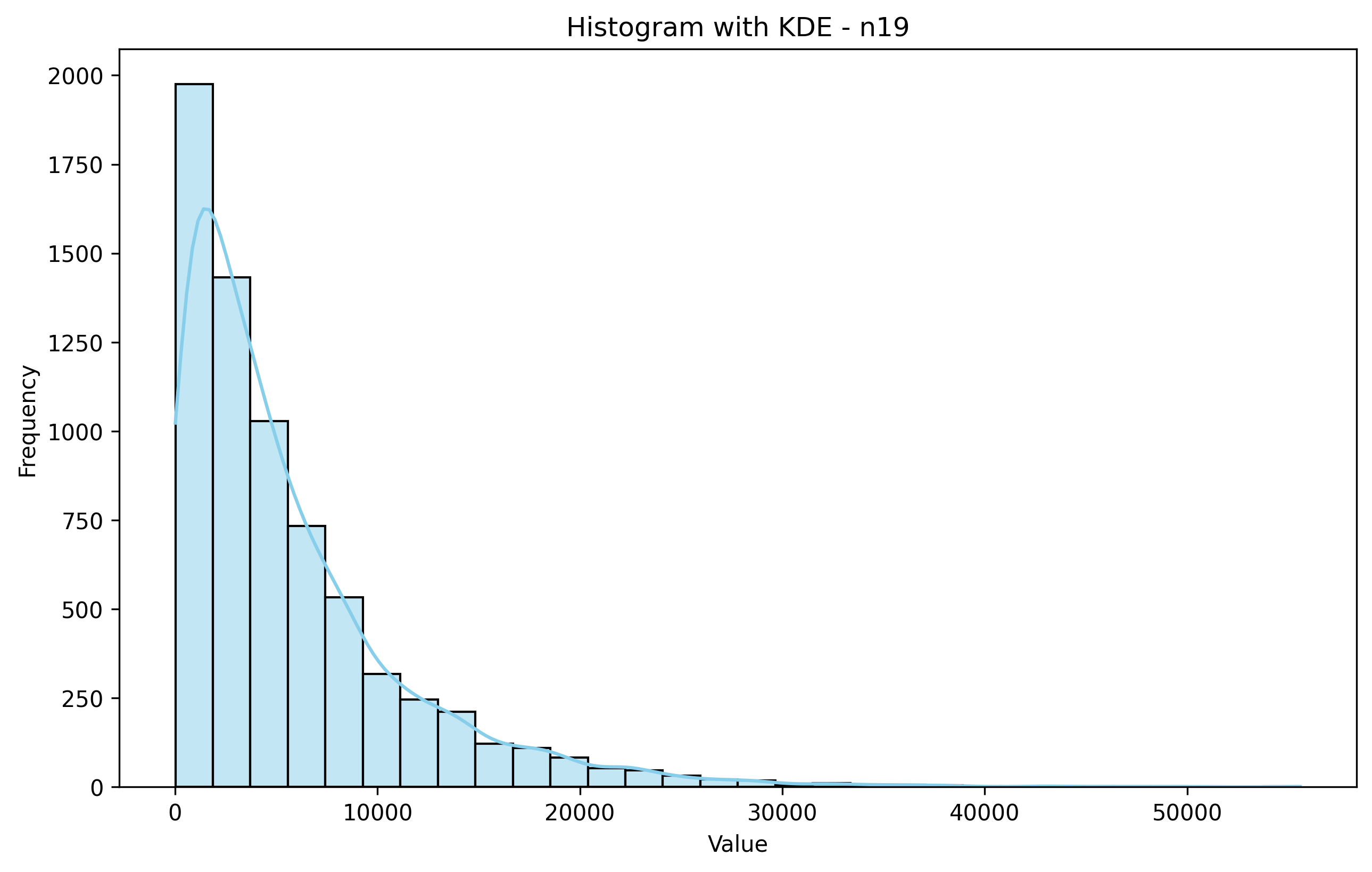}
    \caption{Las Vegas Attempt distribution for n = 22}
\end{subfigure}
\hfill
\
\begin{subfigure}[b]{0.3\textwidth}
    \includegraphics[width=\textwidth]{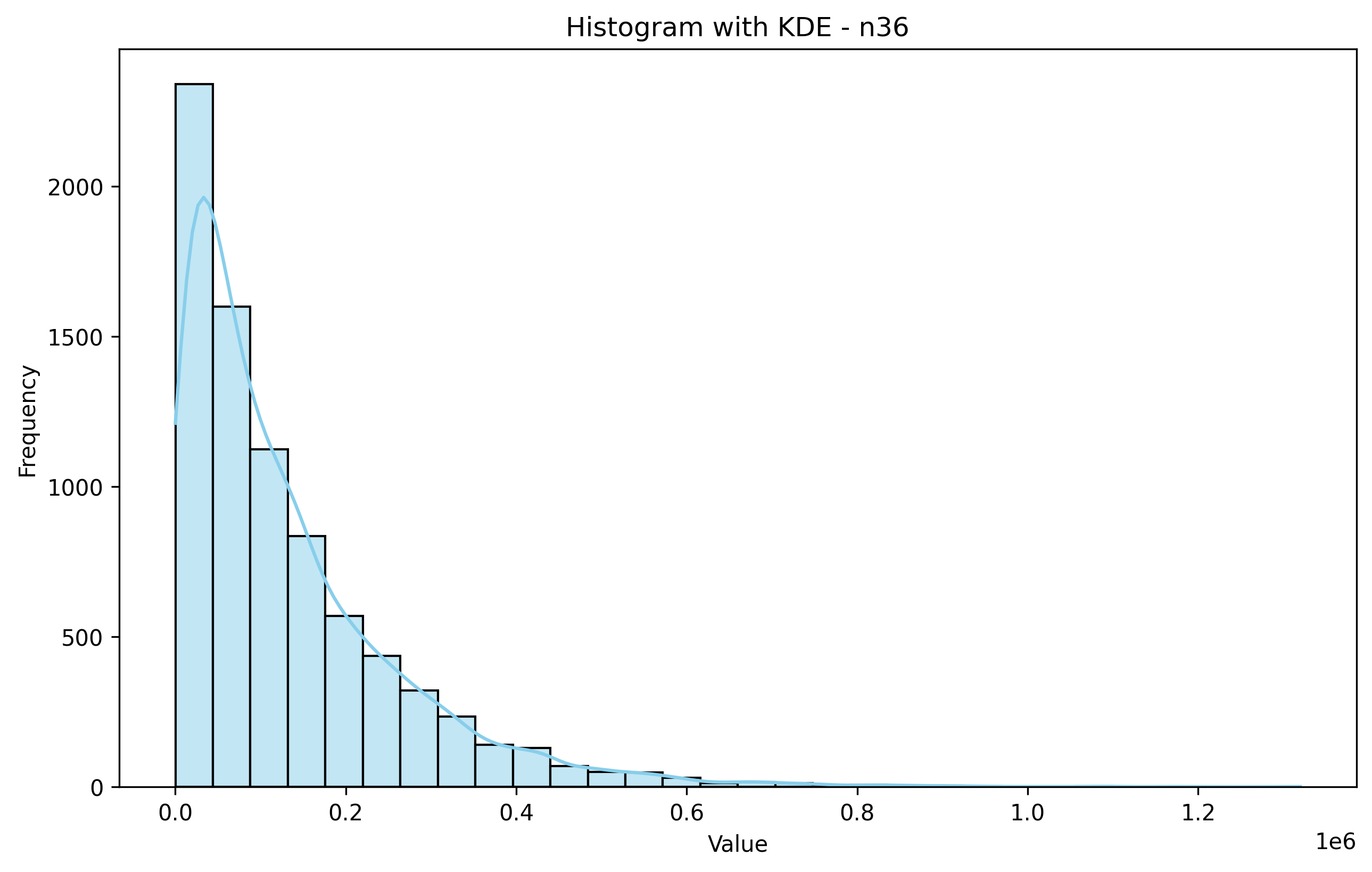}
    \caption{Las Vegas Attempt distribution for n = 36}
\end{subfigure}

\caption{Las Vegas Algorithm Attempt Distributions for  n = 8, 22, 36}
\label{fig:las_vegas_distributions}
\end{figure}

\begin{figure}[H]
    \centering
    \includegraphics[width=0.75\linewidth]{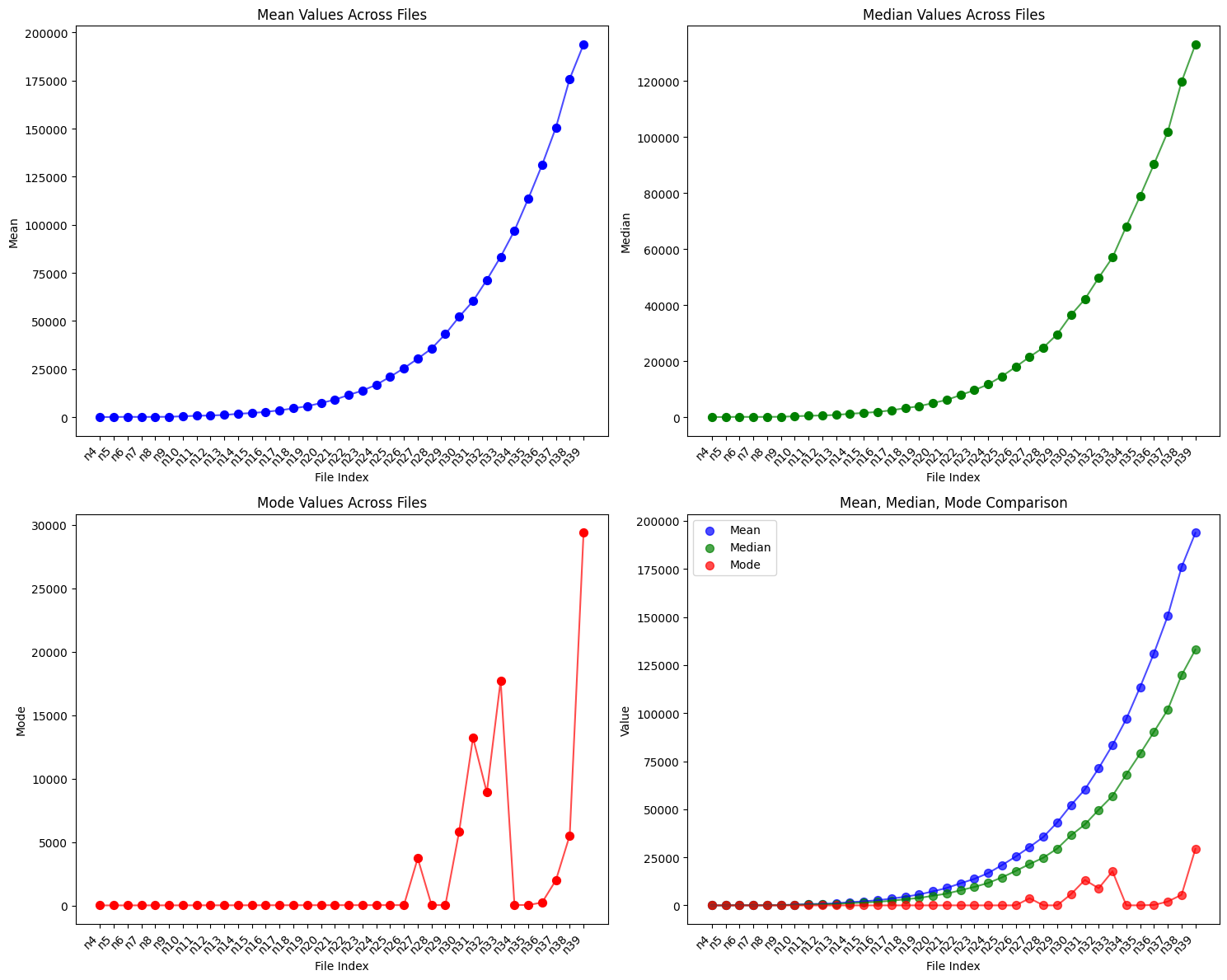}
    \caption{Central Tendencies for Las Vegas}
    \label{fig:LV-central}
\end{figure}

\begin{figure}[H]
    \centering
    \includegraphics[width=0.75\linewidth]{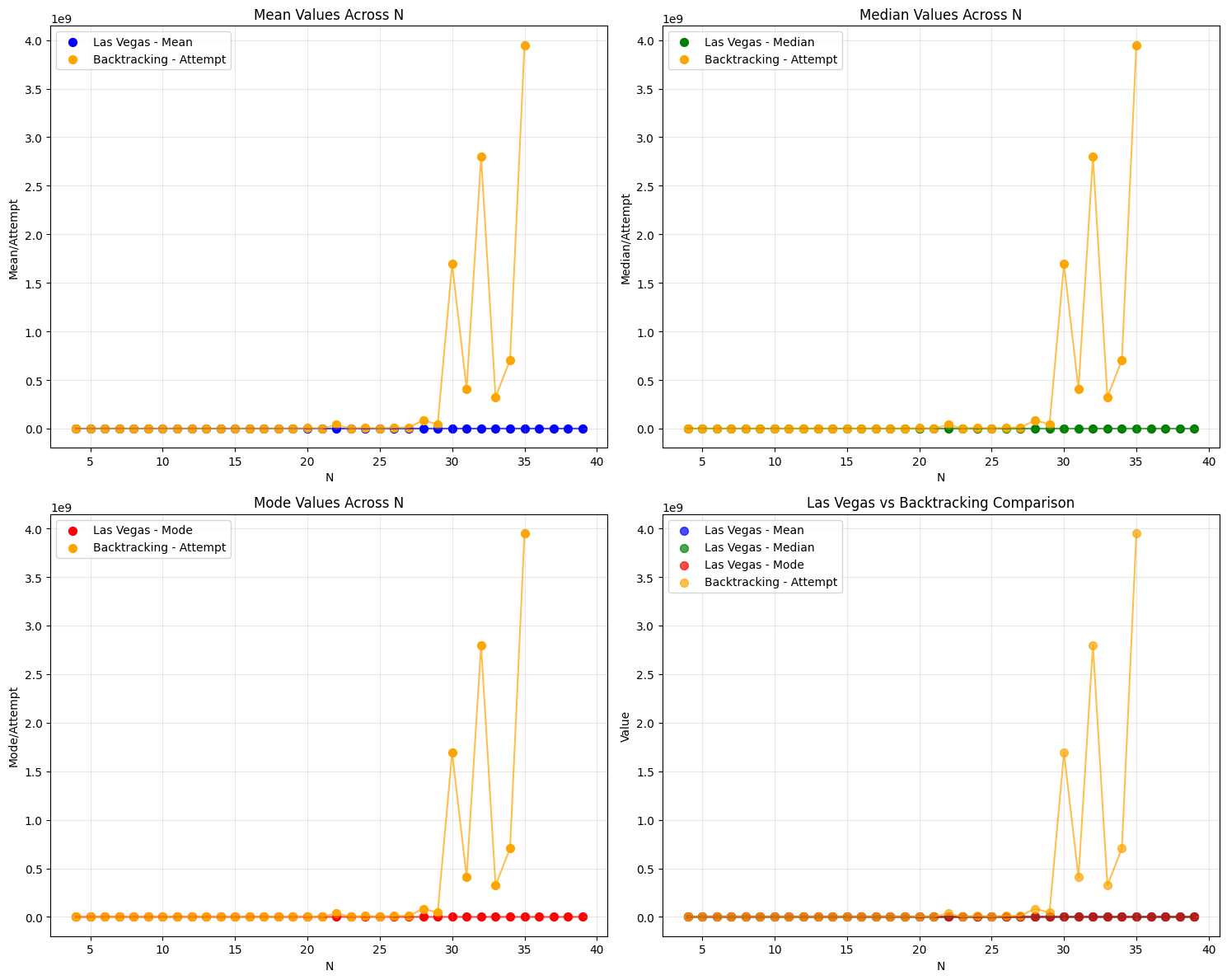}
    \caption{Comparision With Backtracking}
    \label{fig:analysis}
\end{figure}

Figures, \ref{fig:LV-central} and \ref{fig:analysis} comparing mean, median, and mode across different $n$ show that the mean attempts for the probabilistic method increase with n, but at a slower rate than backtracking. The median values follow a similar trend, lying below the mean for all n, as is expected with the positive skew. Lastly, the mode values remain significantly smaller than both the mean and median, suggesting that most runs are fast but occasional long runs increase the average.

These findings confirm that the probabilistic method provides noticeable performance benefits for moderate and large n, particularly when worst-case guarantees are not essential, and solutions can be found with high probability in reasonable time. However, the heavy-tailed distribution means that in rare cases, runtime can be unexpectedly long. This can be mitigated with restart strategies or hybrid approaches.

This research addresses fundamental limitations of existing approaches while opening new avenues for theoretical analysis and practical application. The integration of state space reduction techniques with Las Vegas algorithms for the N-Queens problem represents a novel contribution to both constraint satisfaction methodology and randomized algorithm design.\\

The theoretical significance lies in demonstrating how probabilistic guarantees can be maintained while achieving deterministic space complexity improvements. The practical significance involves providing scalable solutions for large-scale N-Queens instances that exceed the capabilities of traditional backtracking approaches.\\

Future research directions include extending these hybrid techniques to other constraint satisfaction problems, developing theoretical bounds for the expected performance of combined approaches, and investigating the applicability of machine learning techniques for enhancing both state space reduction and probabilistic search guidance.

\subsection{Conclusion And Future Work}
This work presented a hybrid Las Vegas algorithm with state pruning for solving the N-Queens problem, showing that probabilistic search combined with pruning significantly improves efficiency compared to classical backtracking. By dynamically eliminating invalid placements during the random assignment phase, the algorithm achieves faster convergence while preserving correctness. The results demonstrate its promise as a practical alternative where timely solutions are preferred over completeness.\\

Future directions are threefold. First, although the present contribution focuses on N-Queens, the approach can be framed more broadly as a hybrid stochastic–deterministic strategy for constraint satisfaction problems (CSPs), with potential applications in scheduling, graph coloring, and resource allocation. Second, while correctness was established through loop invariants, a deeper probabilistic runtime analysis is needed to formalize expected performance bounds. Finally, extending the experimental scope to include competitive baselines such as the min-conflicts heuristic, SAT/CP-SAT solvers, and metaheuristic approaches will provide a stronger comparative foundation. These extensions would not only strengthen the theoretical contribution but also clarify the broader applicability of the proposed method as a scalable tool for CSPs.

\section{Appendix}
\begin{appendices}

\subsection{Additional Visualizations} \label{app:figures}

The Histogram with KDE plot for all n

\begin{figure}[H]
\centering
\begin{subfigure}[b]{0.3\textwidth}
    \includegraphics[width=\textwidth]{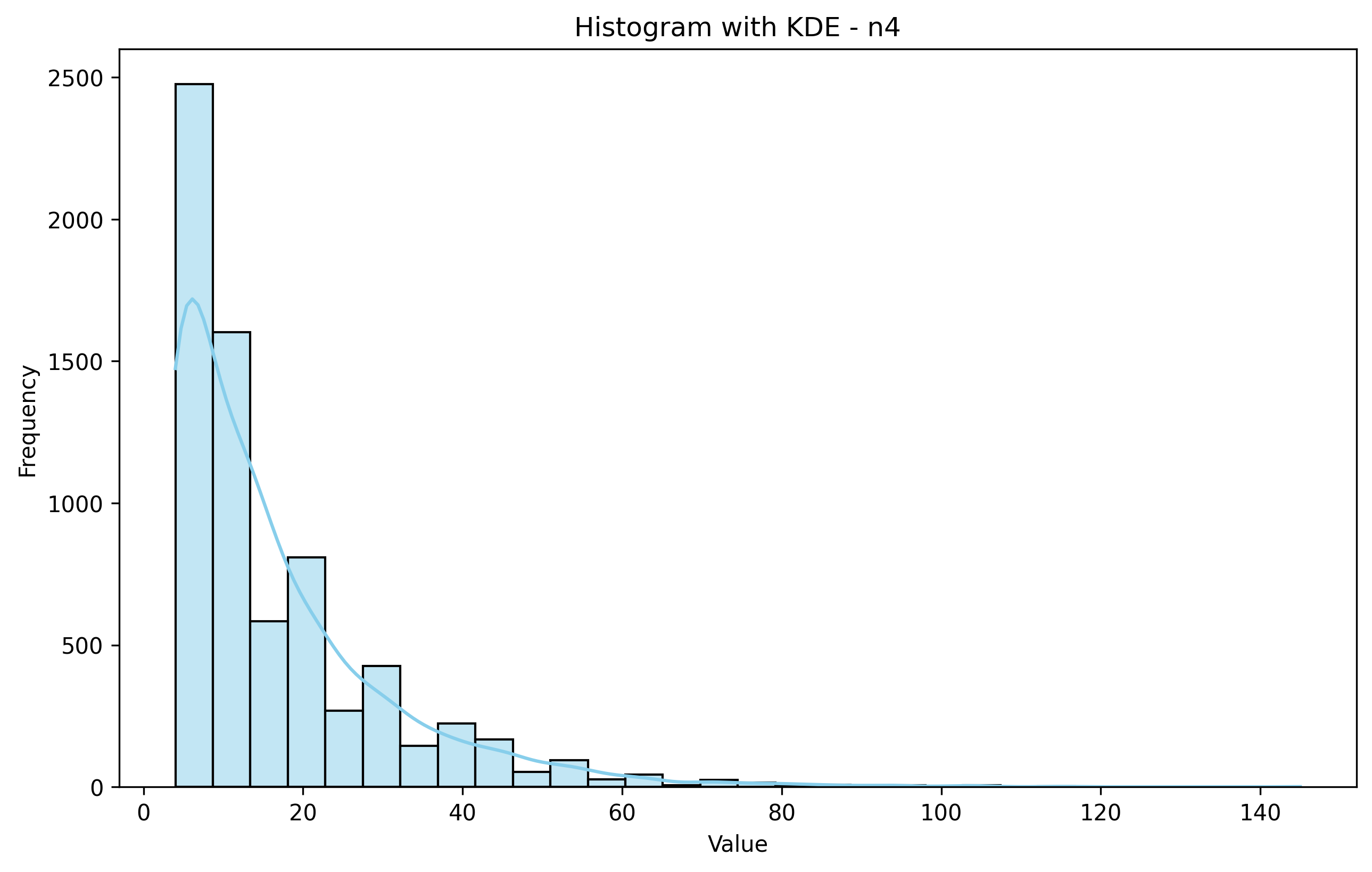}
    \caption{Las Vegas Attempt distribution for n = 4}
\end{subfigure}
\hfill
\begin{subfigure}[b]{0.3\textwidth}
    \includegraphics[width=\textwidth]{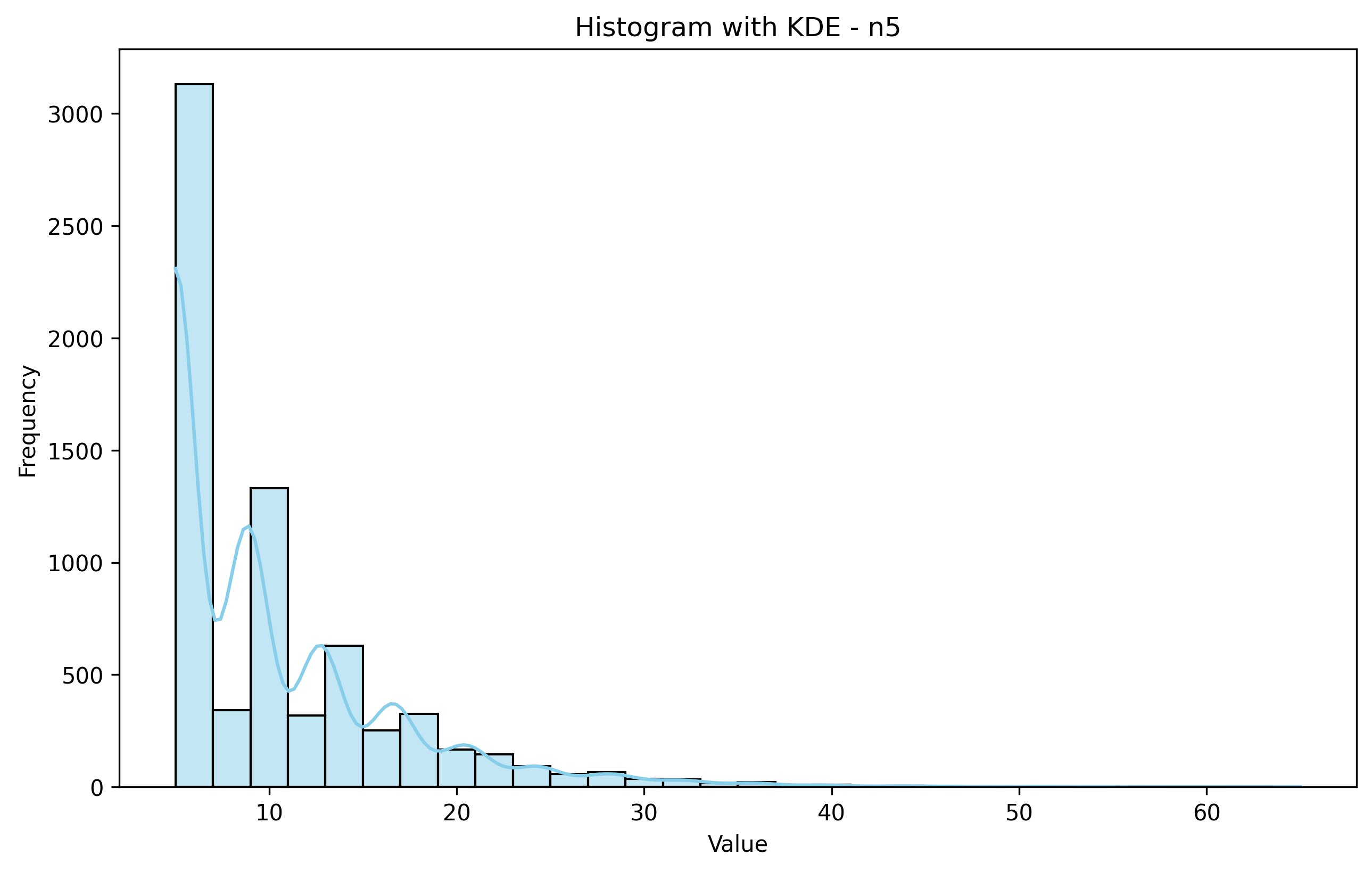}
    \caption{Las Vegas Attempt distribution for n = 5}
\end{subfigure}
\hfill
\begin{subfigure}[b]{0.3\textwidth}
    \includegraphics[width=\textwidth]{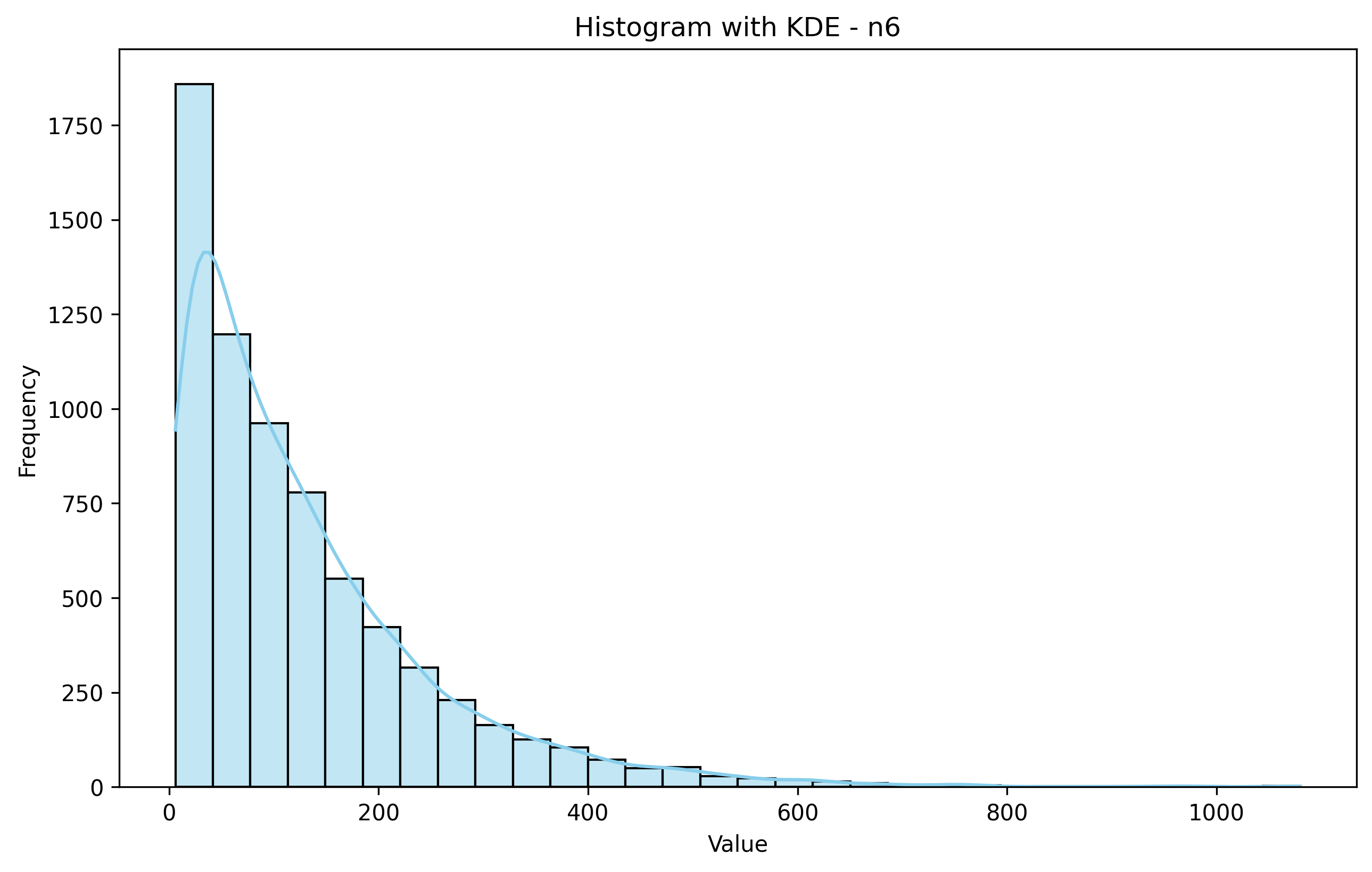}
    \caption{Las Vegas Attempt distribution for n = 6}
\end{subfigure}
\begin{subfigure}[b]{0.3\textwidth}
    \includegraphics[width=\textwidth]{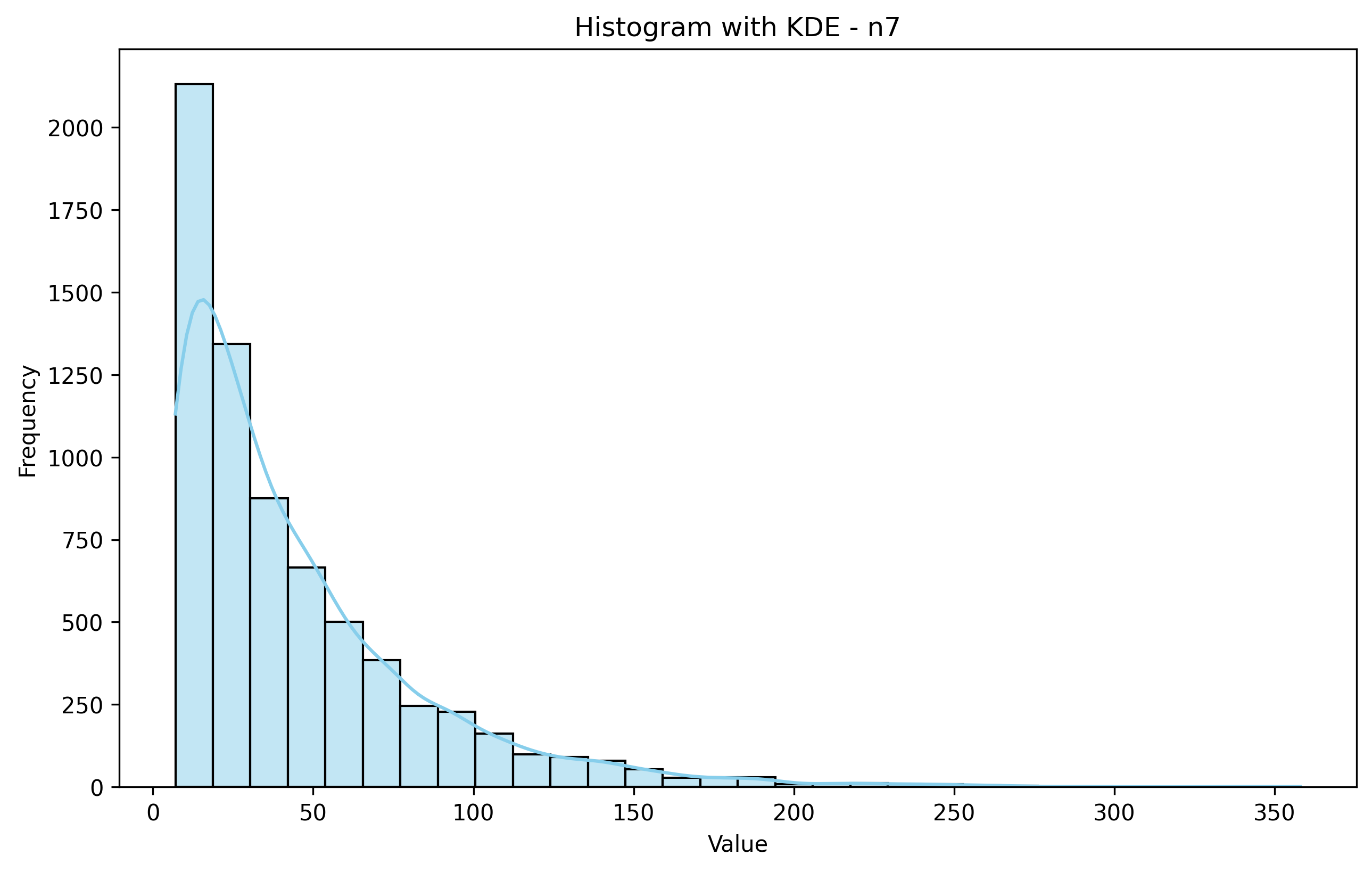}
    \caption{Las Vegas Attempt distribution for n = 7}
\end{subfigure}
\hfill
\begin{subfigure}[b]{0.3\textwidth}
    \includegraphics[width=\textwidth]{n8_histogram.png}
    \caption{Las Vegas Attempt distribution for n = 8}
\end{subfigure}
\hfill
\begin{subfigure}[b]{0.3\textwidth}
    \includegraphics[width=\textwidth]{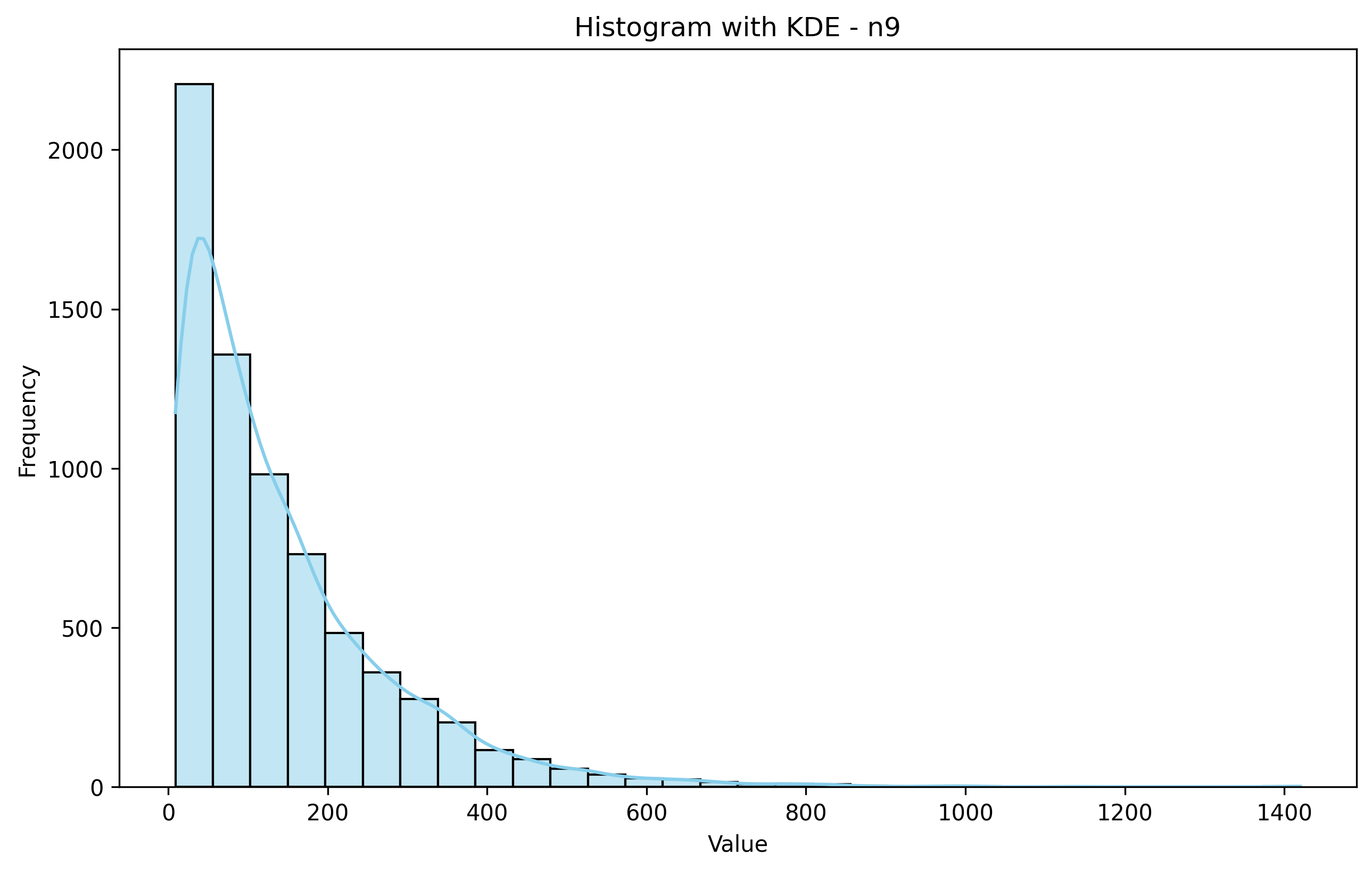}
    \caption{Las Vegas Attempt distribution for n = 9}
\end{subfigure}

\begin{subfigure}[b]{0.3\textwidth}
    \includegraphics[width=\textwidth]{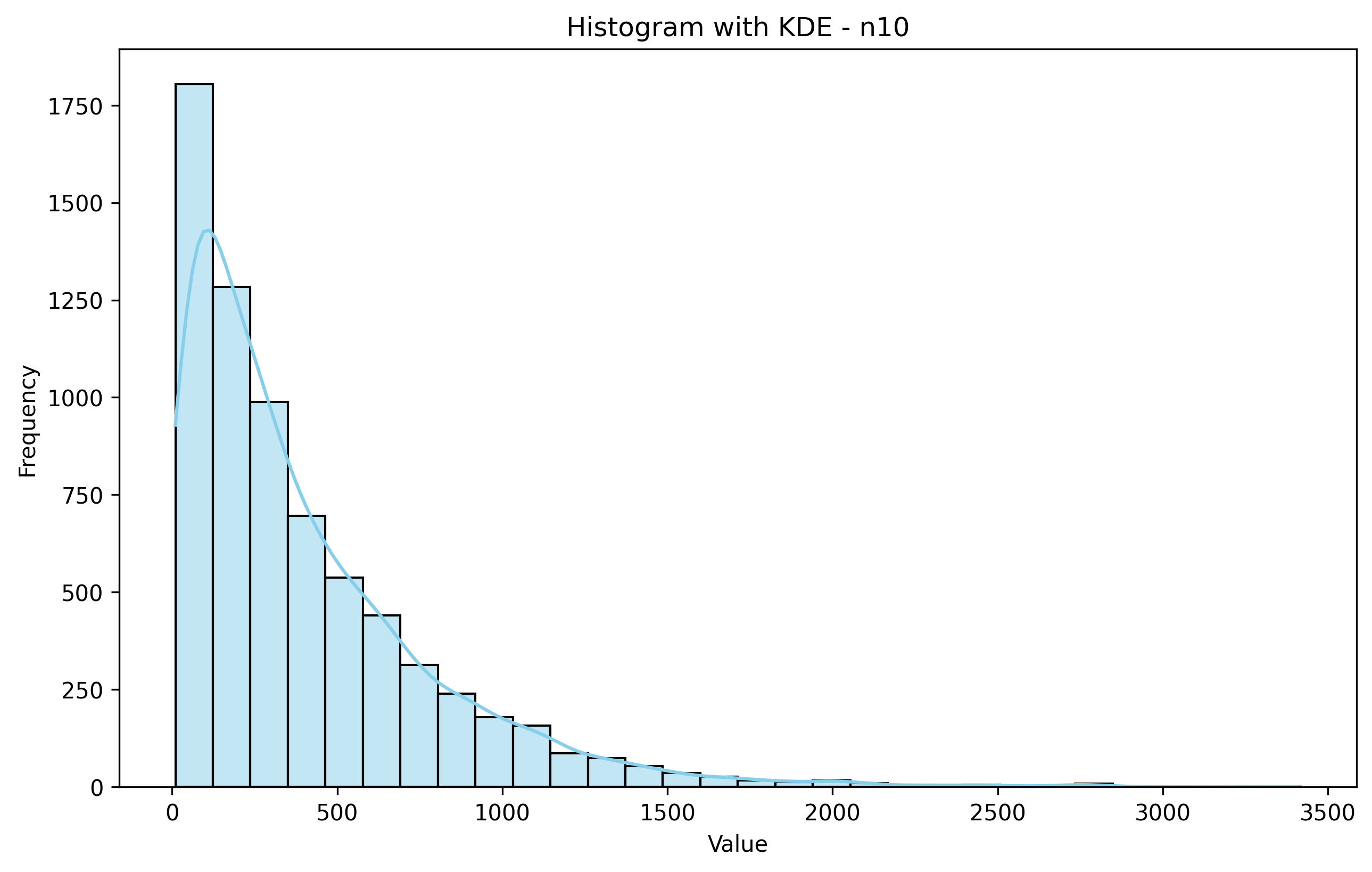}
    \caption{Las Vegas Attempt distribution for n = 10}
\end{subfigure}
\hfill
\begin{subfigure}[b]{0.3\textwidth}
    \includegraphics[width=\textwidth]{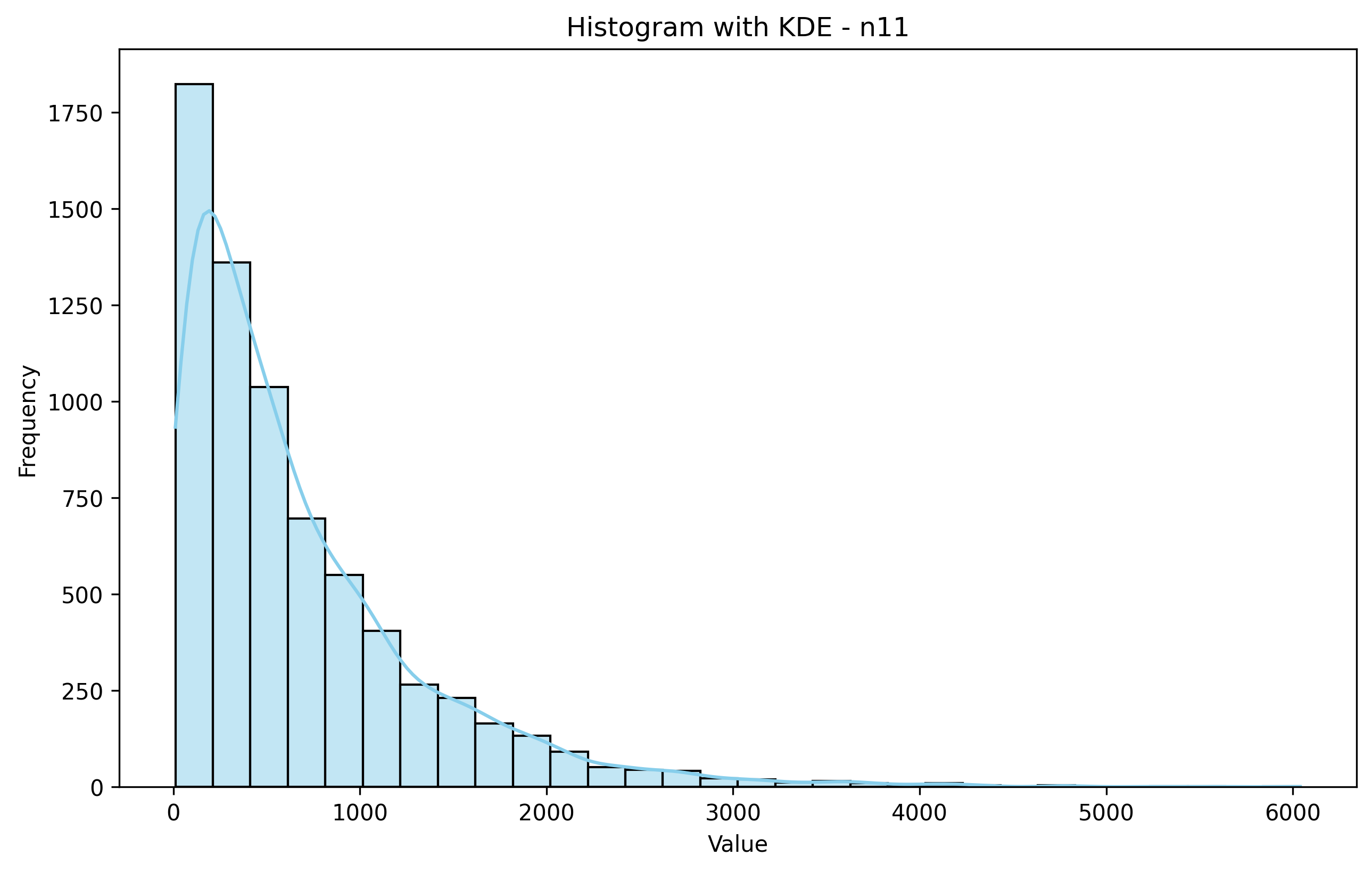}
    \caption{Las Vegas Attempt distribution for n = 11}
\end{subfigure}
\hfill
\begin{subfigure}[b]{0.3\textwidth}
    \includegraphics[width=\textwidth]{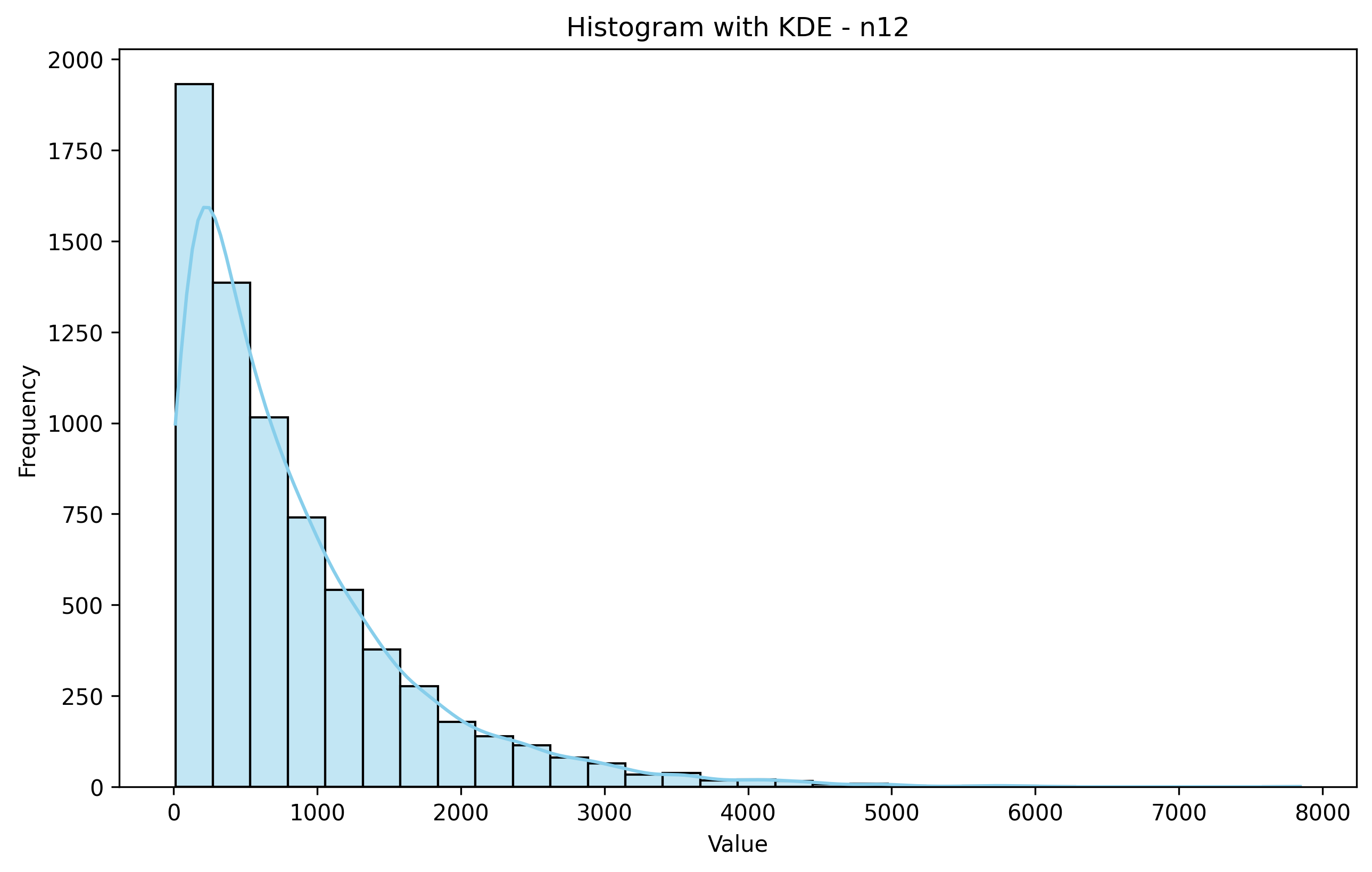}
    \caption{Las Vegas Attempt distribution for n = 12}
\end{subfigure}

\begin{subfigure}[b]{0.3\textwidth}
    \includegraphics[width=\textwidth]{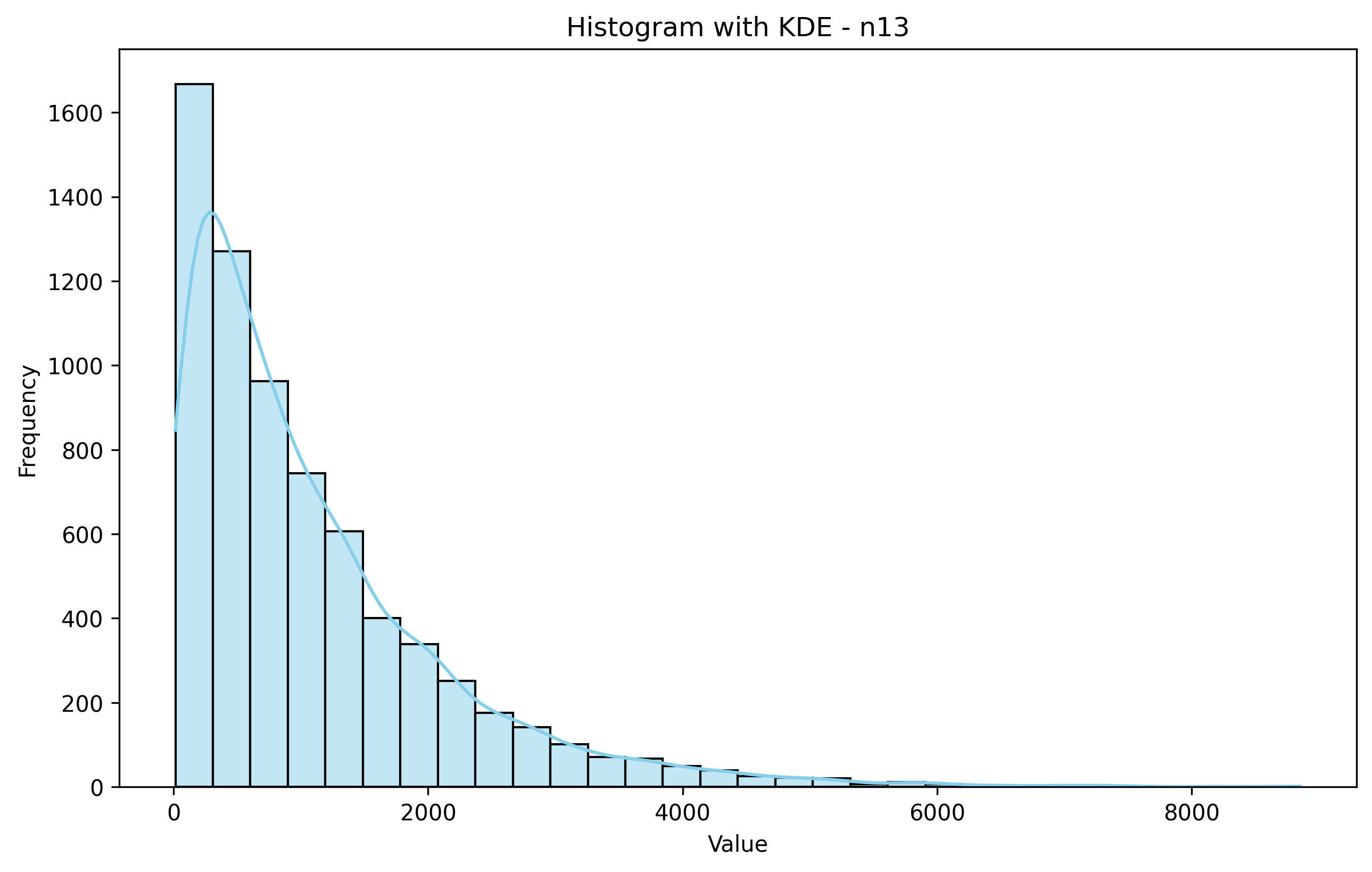}
    \caption{Las Vegas Attempt distribution for n = 13}
\end{subfigure}
\hfill
\begin{subfigure}[b]{0.3\textwidth}
    \includegraphics[width=\textwidth]{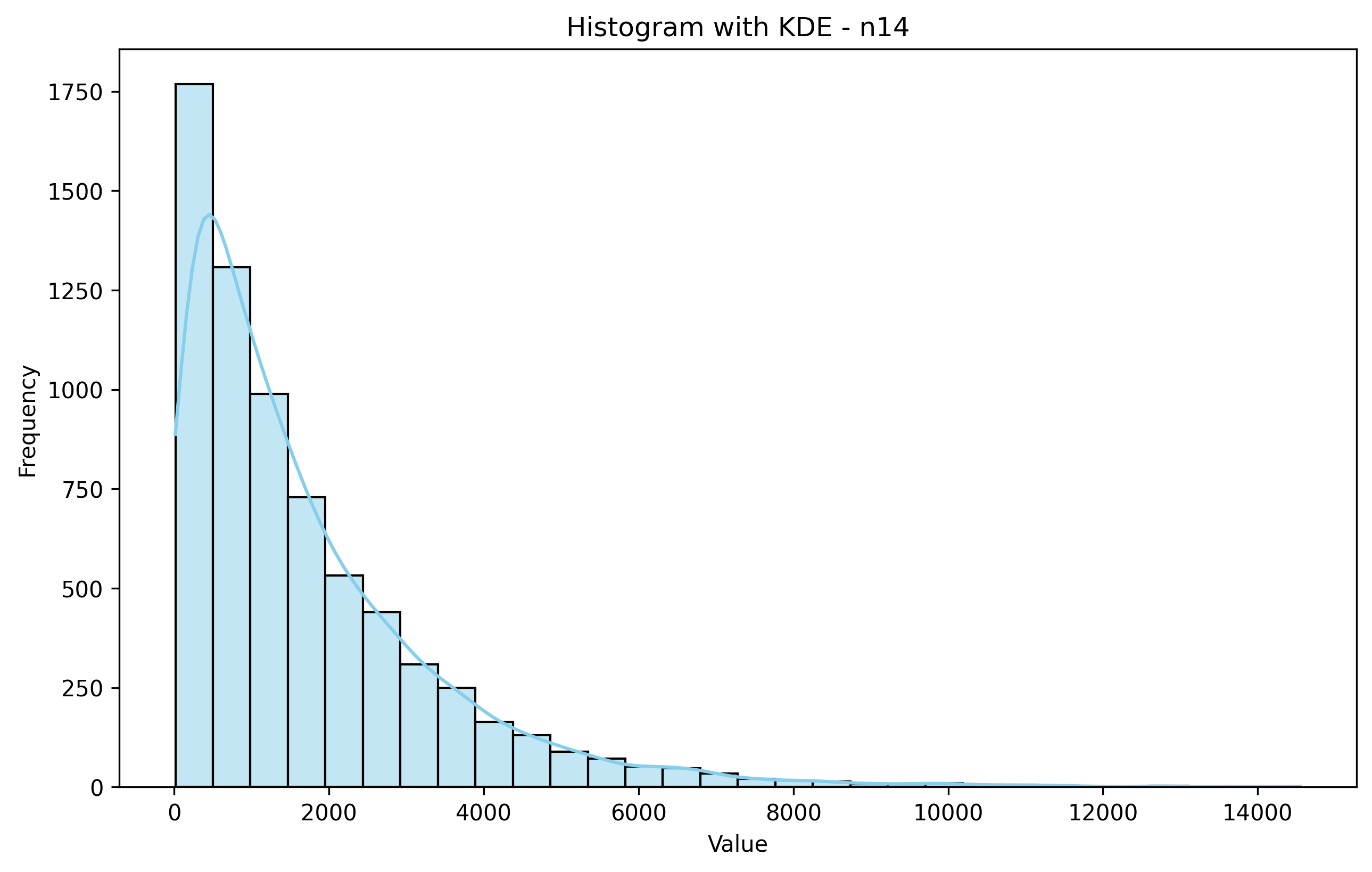}
    \caption{Las Vegas Attempt distribution for n = 14}
\end{subfigure}
\hfill
\begin{subfigure}[b]{0.3\textwidth}
    \includegraphics[width=\textwidth]{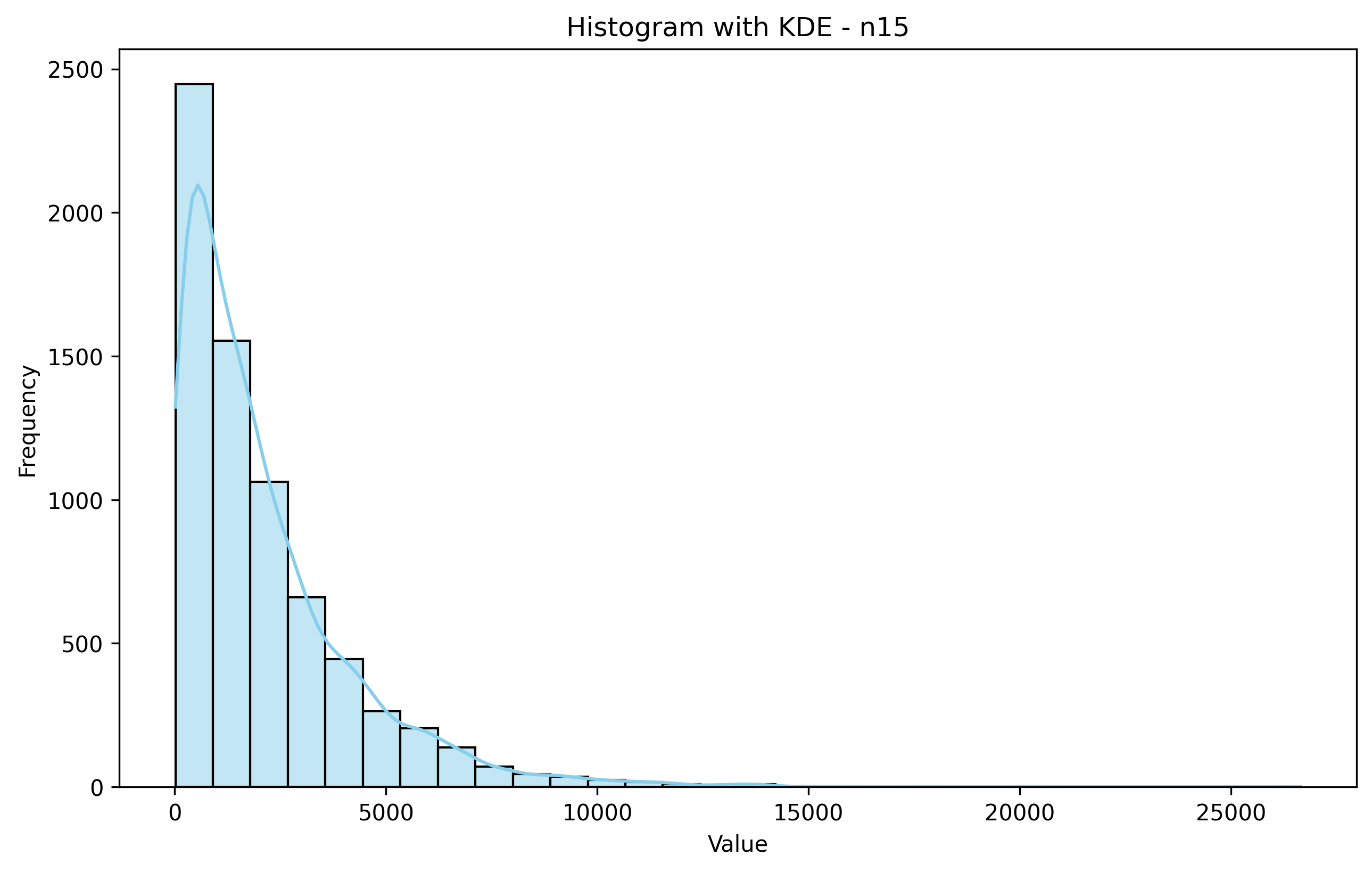}
    \caption{Las Vegas Attempt distribution for n = 15}
\end{subfigure}

\begin{subfigure}[b]{0.3\textwidth}
    \includegraphics[width=\textwidth]{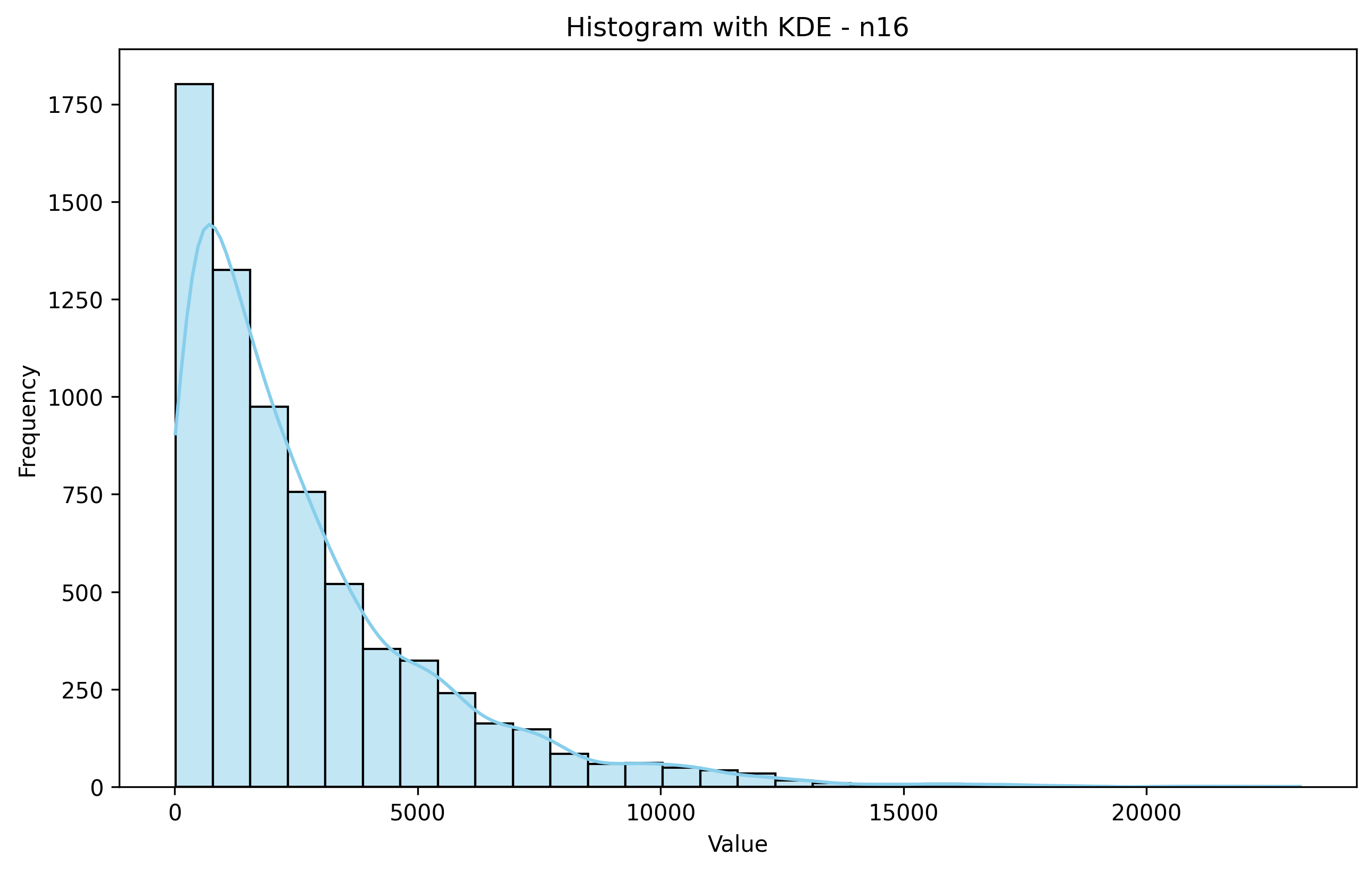}
    \caption{Las Vegas Attempt distribution for n = 16}
\end{subfigure}
\hfill
\begin{subfigure}[b]{0.3\textwidth}
    \includegraphics[width=\textwidth]{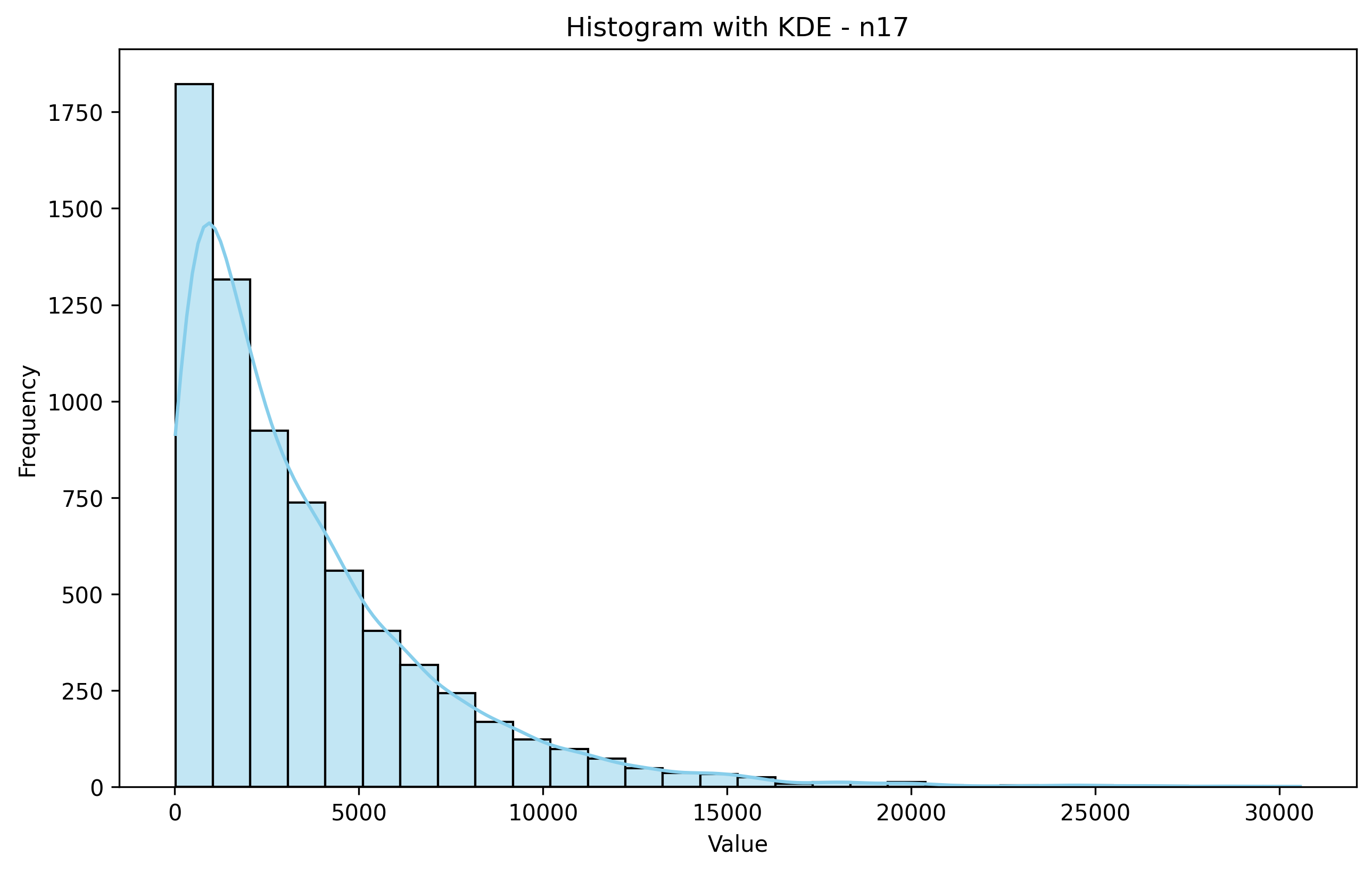}
    \caption{Las Vegas Attempt distribution for n = 17}
\end{subfigure}
\hfill
\begin{subfigure}[b]{0.3\textwidth}
    \includegraphics[width=\textwidth]{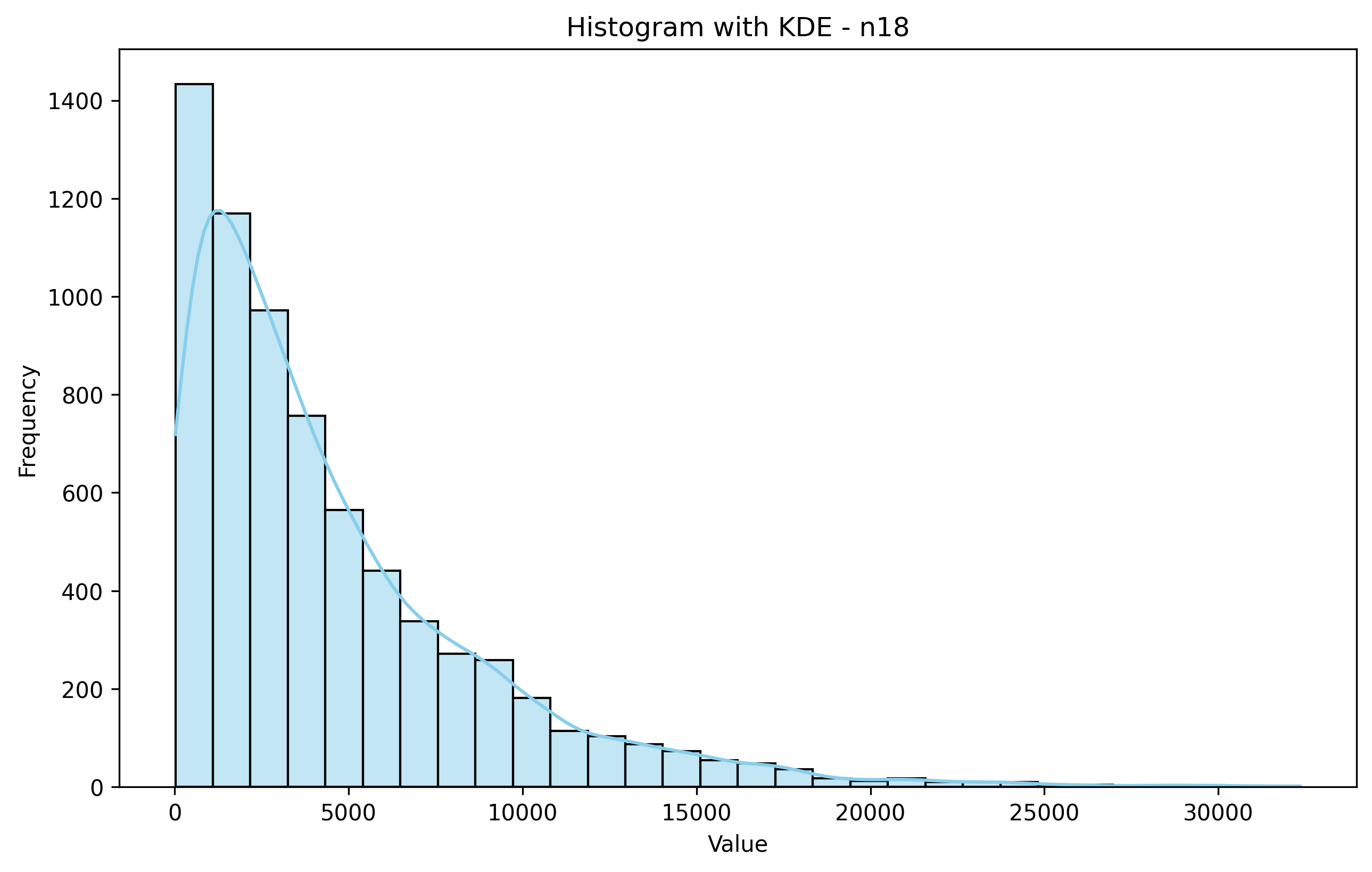}
    \caption{Las Vegas Attempt distribution for n = 18}
\end{subfigure}

\begin{subfigure}[b]{0.3\textwidth}
    \includegraphics[width=\textwidth]{n19_histogram.png}
    \caption{Las Vegas Attempt distribution for n = 19}
\end{subfigure}
\hfill
\begin{subfigure}[b]{0.3\textwidth}
    \includegraphics[width=\textwidth]{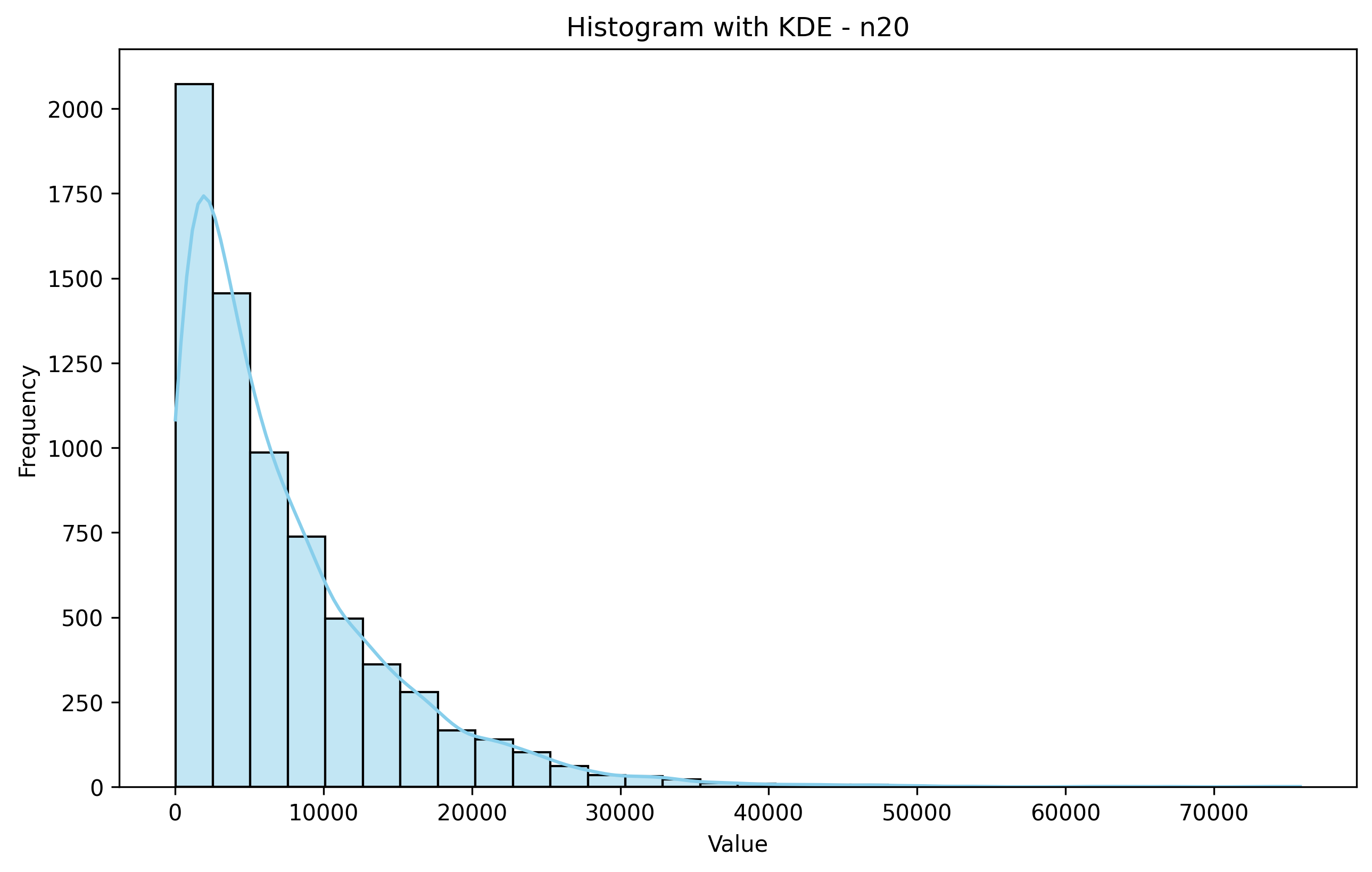}
    \caption{Las Vegas Attempt distribution for n = 20}
\end{subfigure}
\hfill
\begin{subfigure}[b]{0.3\textwidth}
    \includegraphics[width=\textwidth]{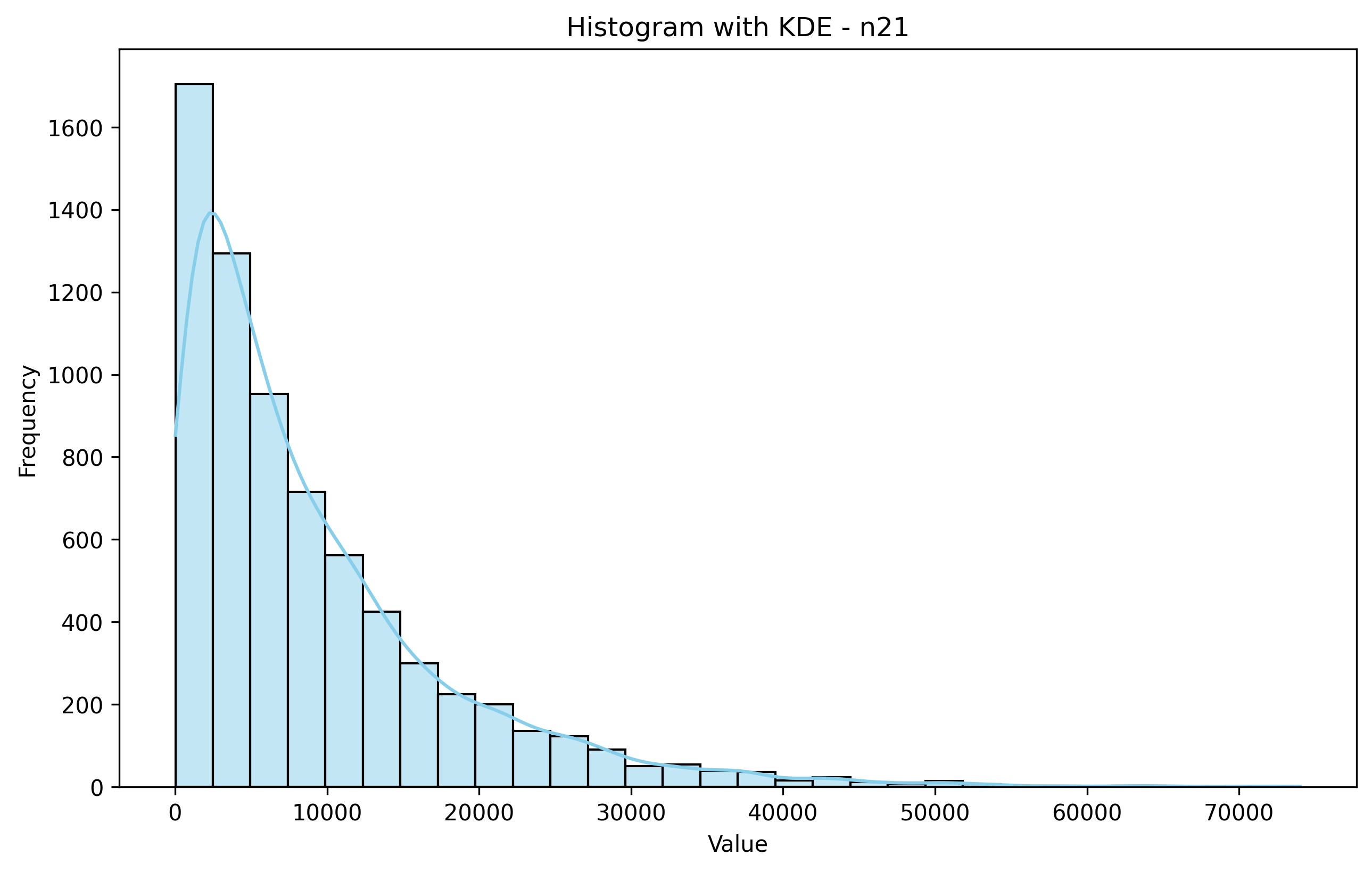}
    \caption{Las Vegas Attempt distribution for n = 21}
\end{subfigure}
\end{figure}

\begin{figure}[H]
\centering
\begin{subfigure}[b]{0.3\textwidth}
    \includegraphics[width=\textwidth]{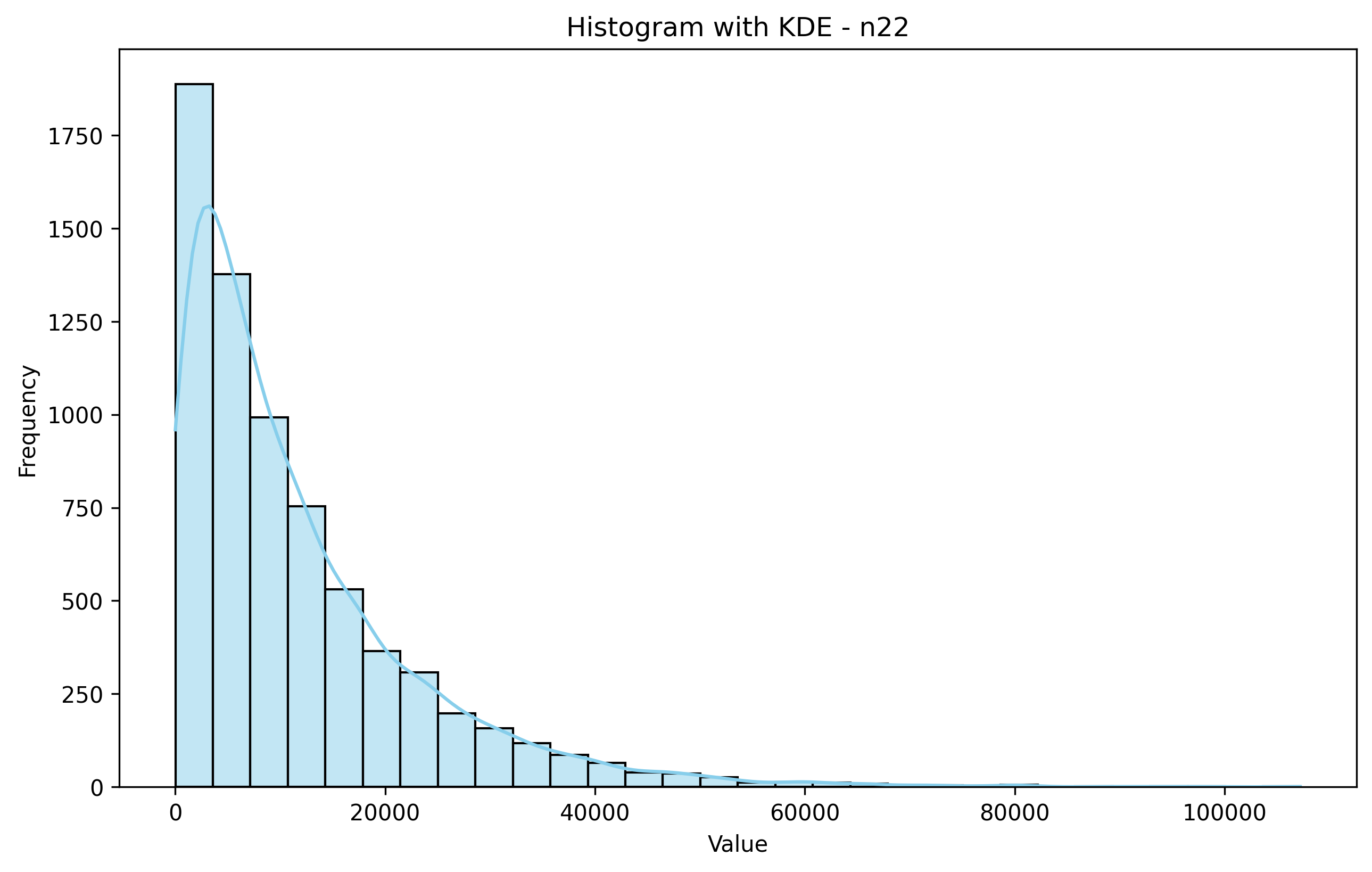}
    \caption{Las Vegas Attempt distribution for n = 22}
\end{subfigure}
\hfill
\begin{subfigure}[b]{0.3\textwidth}
    \includegraphics[width=\textwidth]{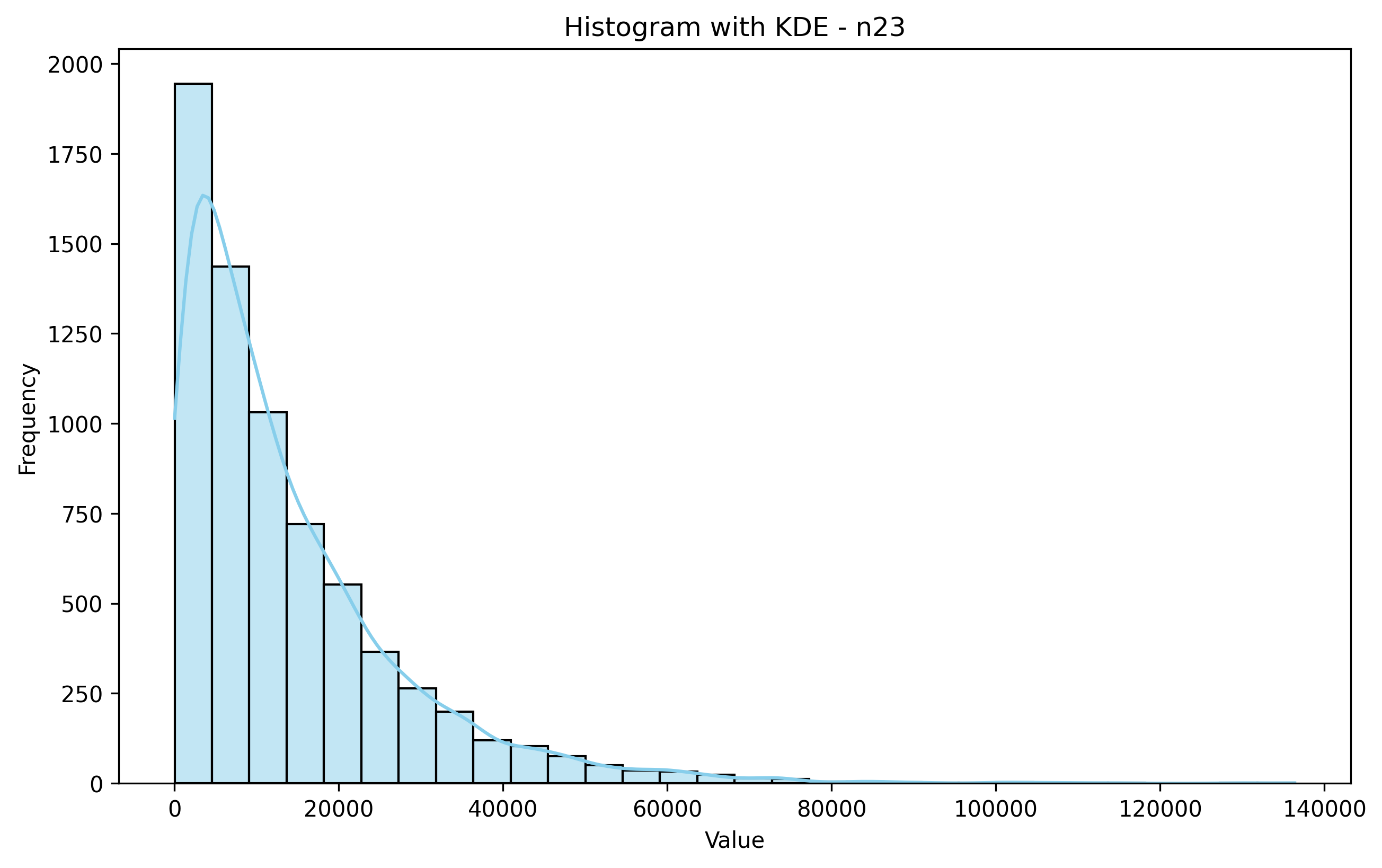}
    \caption{Las Vegas Attempt distribution for n = 23}
\end{subfigure}
\hfill
\begin{subfigure}[b]{0.3\textwidth}
    \includegraphics[width=\textwidth]{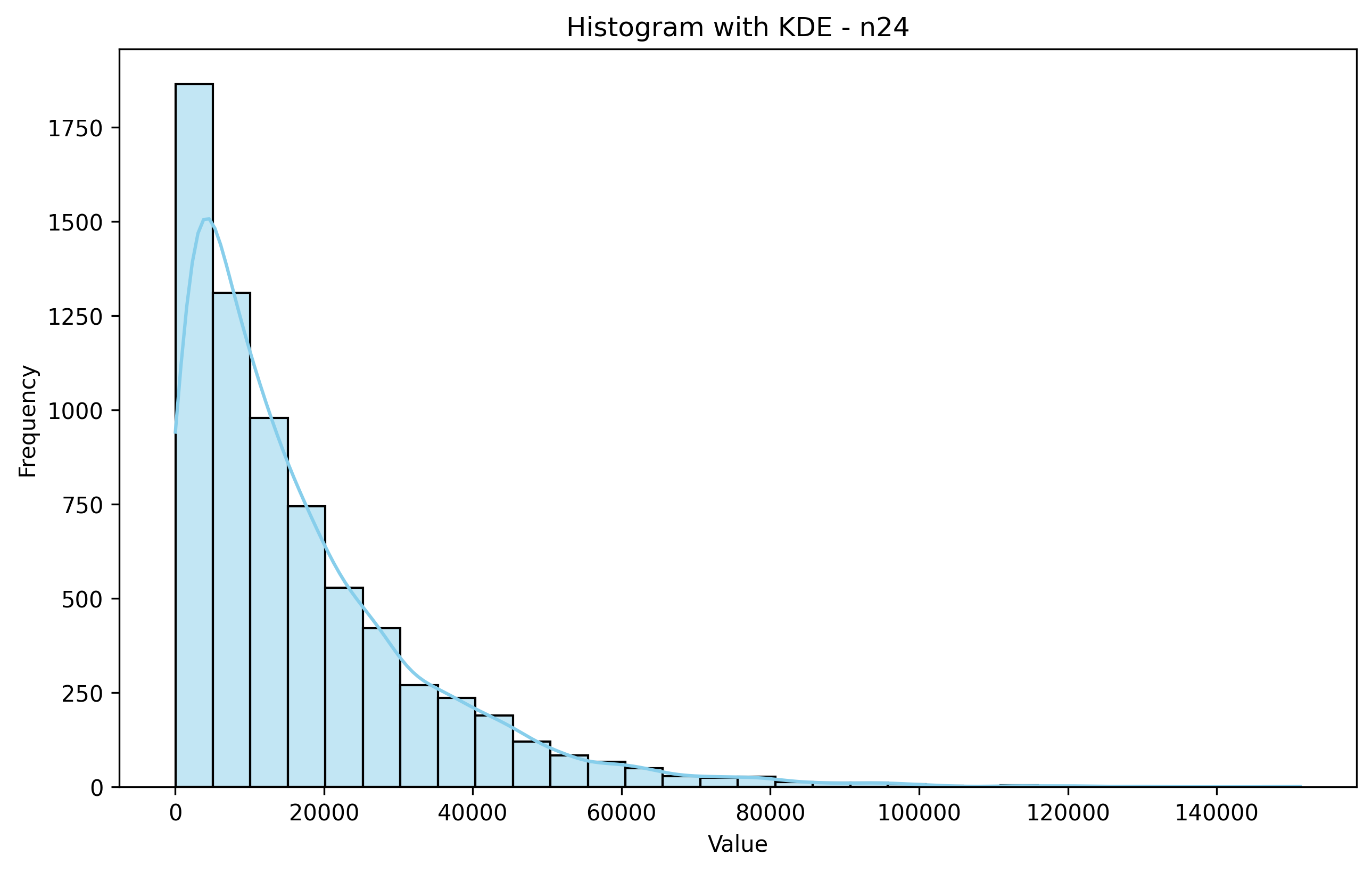}
    \caption{Las Vegas Attempt distribution for n = 24}
\end{subfigure}

\begin{subfigure}[b]{0.3\textwidth}
    \includegraphics[width=\textwidth]{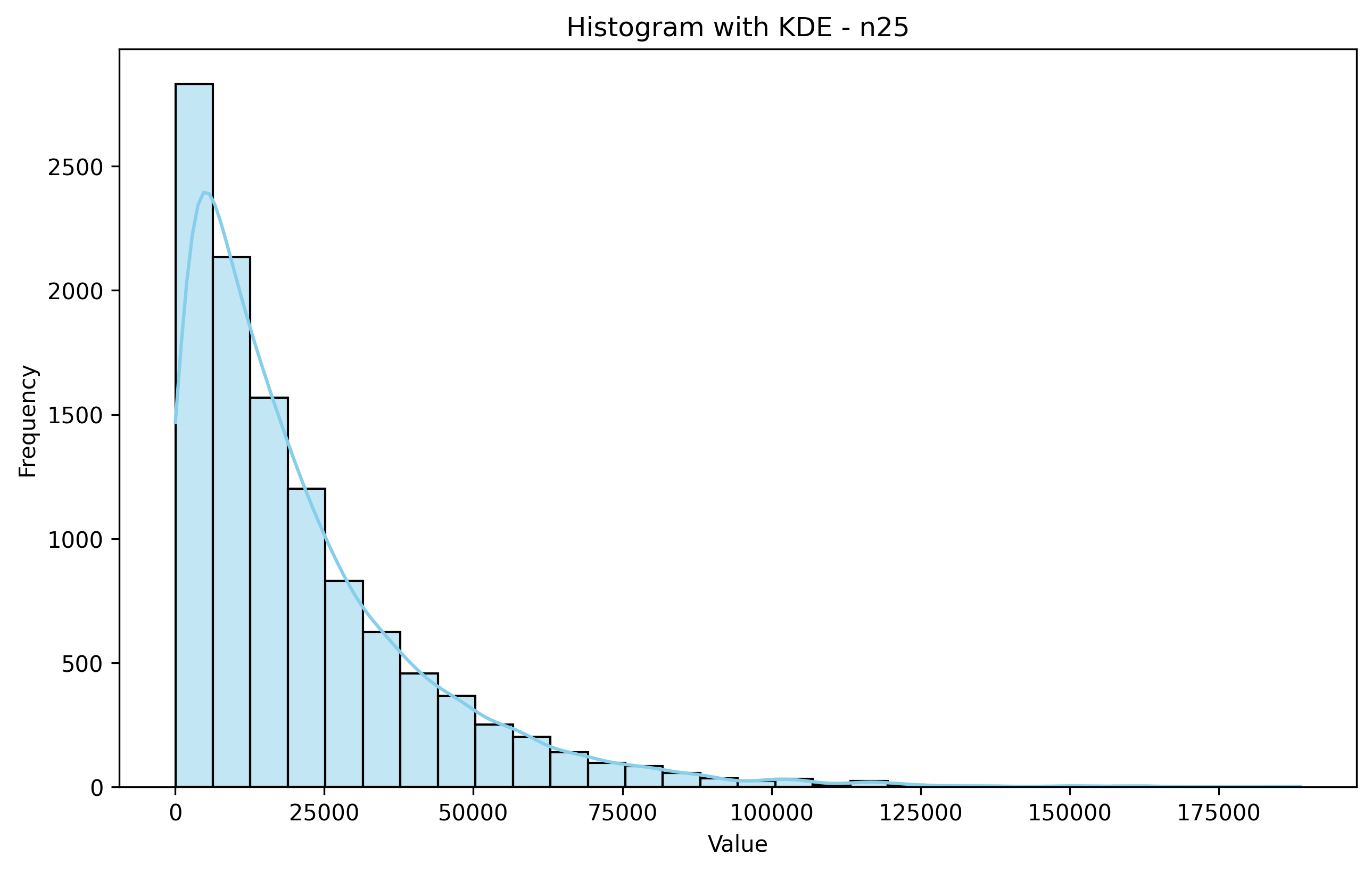}
    \caption{Las Vegas Attempt distribution for n = 25}
\end{subfigure}
\hfill
\begin{subfigure}[b]{0.3\textwidth}
    \includegraphics[width=\textwidth]{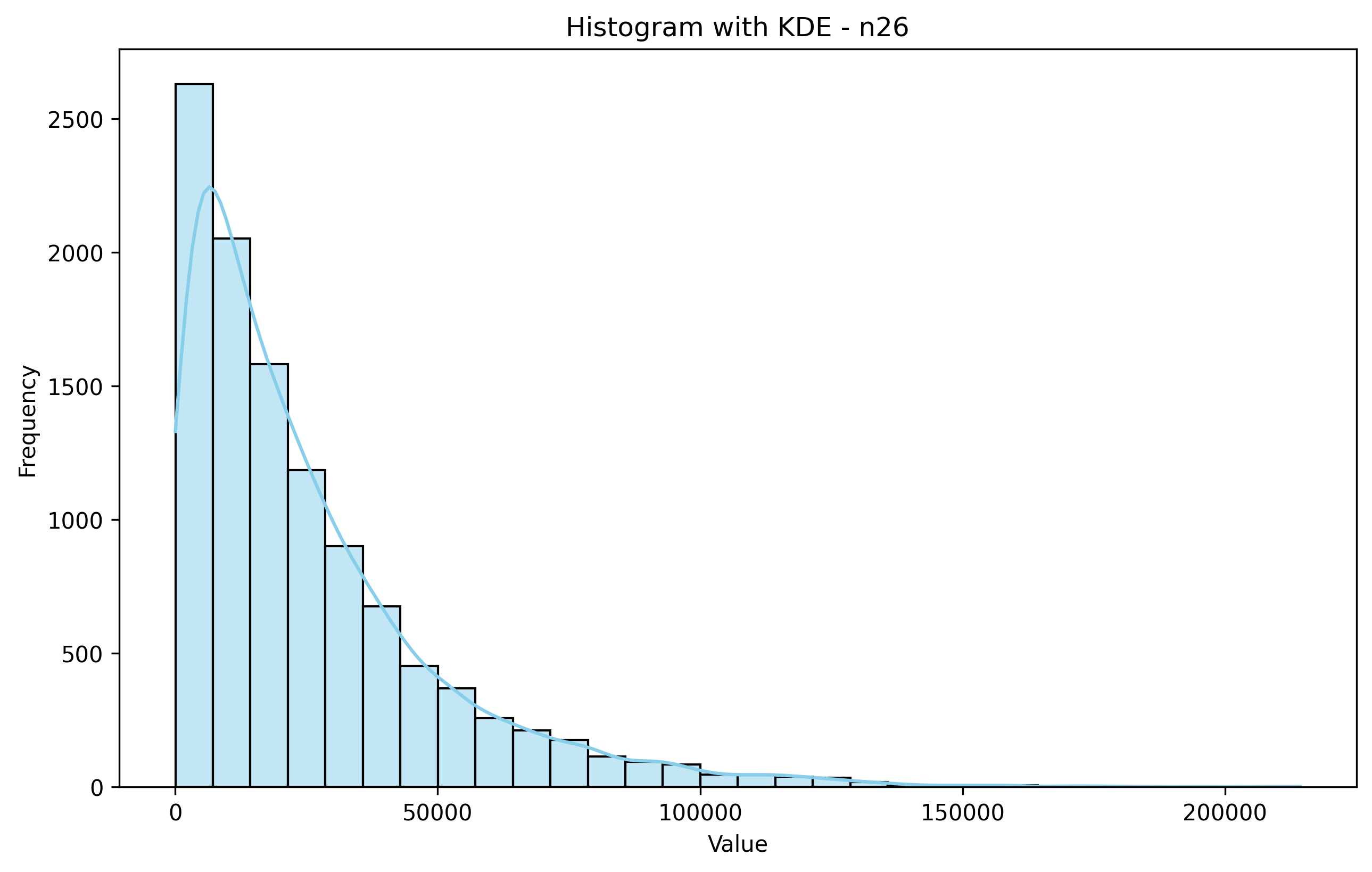}
    \caption{Las Vegas Attempt distribution for n = 26}
\end{subfigure}
\hfill
\begin{subfigure}[b]{0.3\textwidth}
    \includegraphics[width=\textwidth]{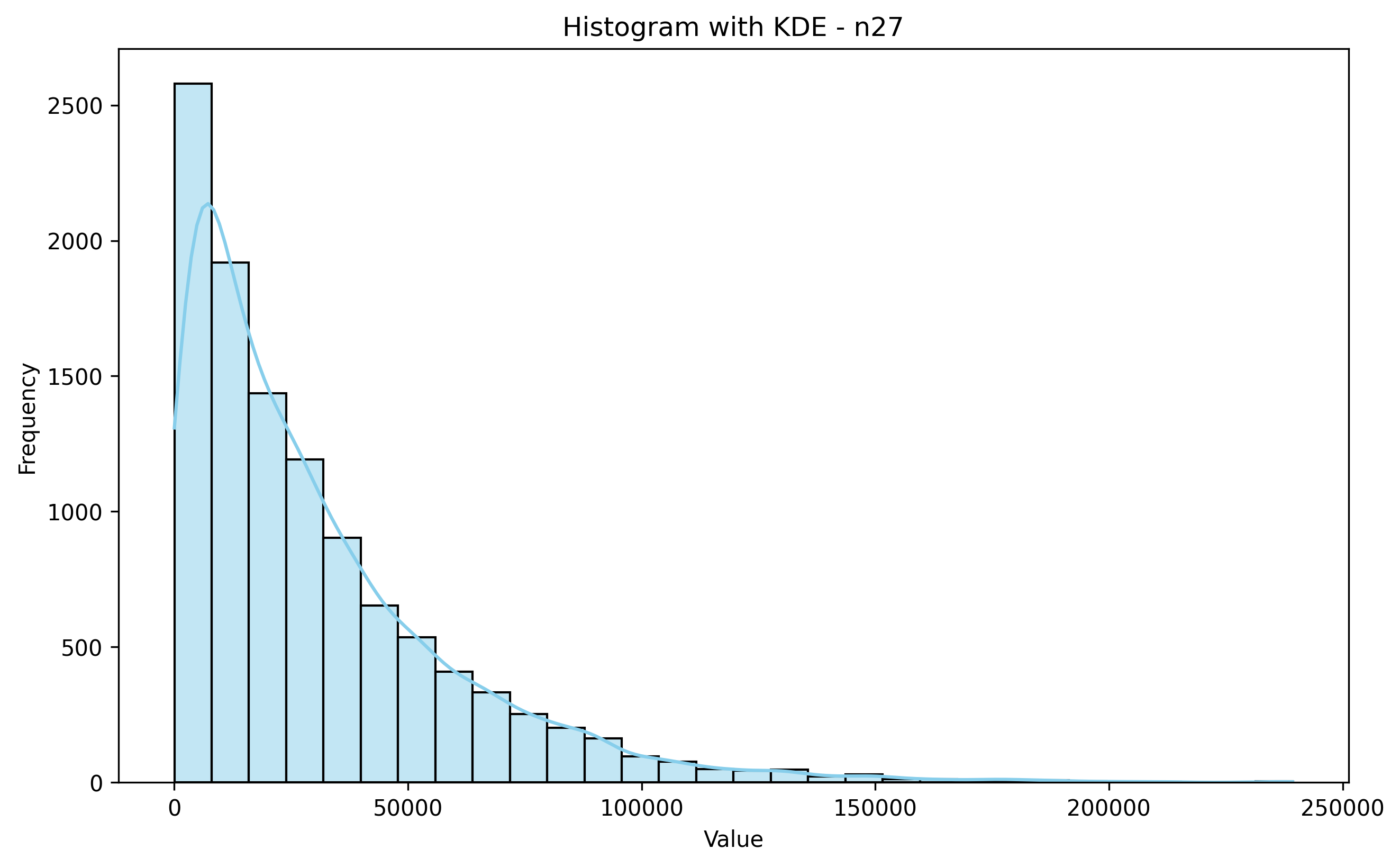}
    \caption{Las Vegas Attempt distribution for n = 27}
\end{subfigure}

\begin{subfigure}[b]{0.3\textwidth}
    \includegraphics[width=\textwidth]{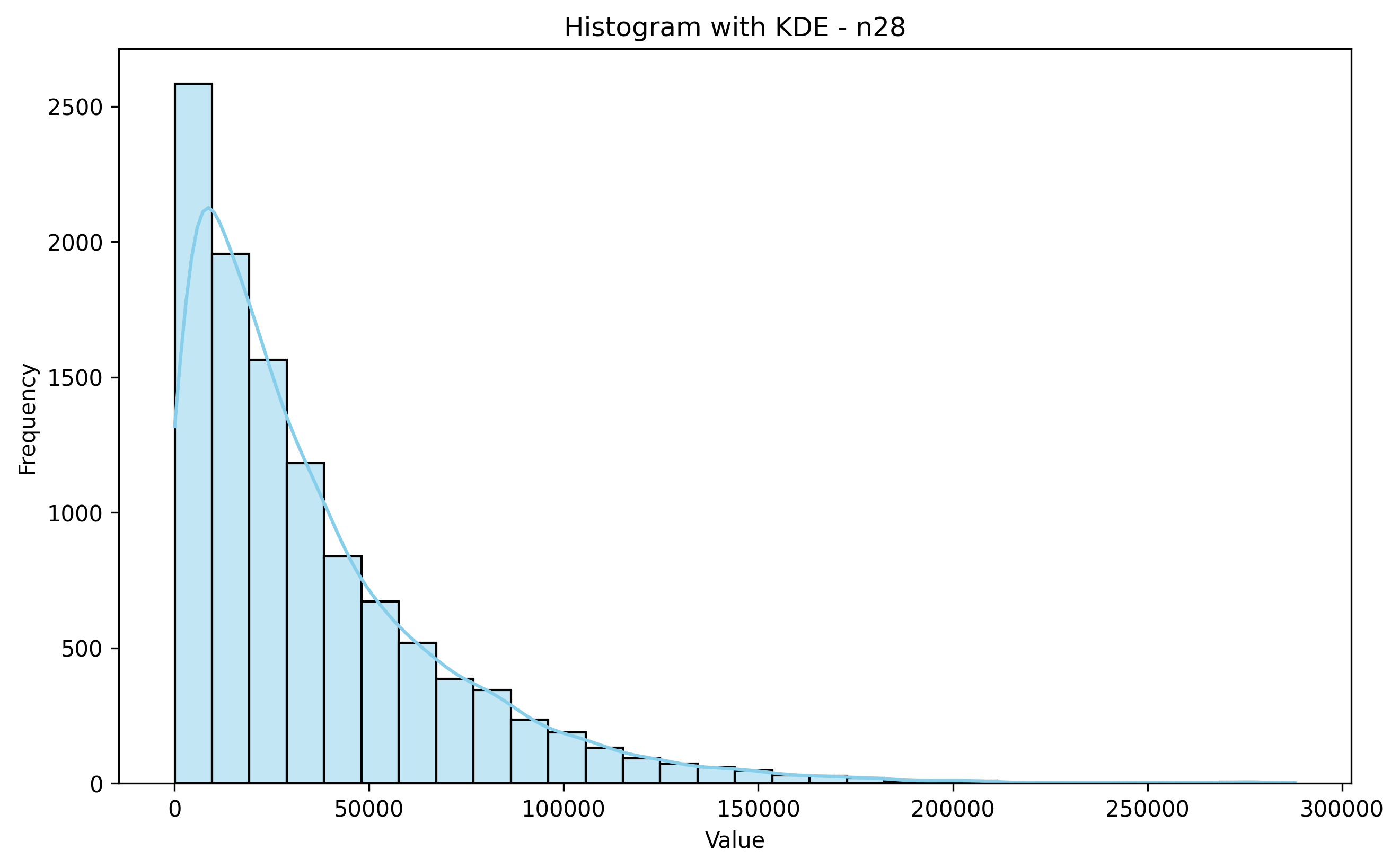}
    \caption{Las Vegas Attempt distribution for n = 28}
\end{subfigure}
\hfill
\begin{subfigure}[b]{0.3\textwidth}
    \includegraphics[width=\textwidth]{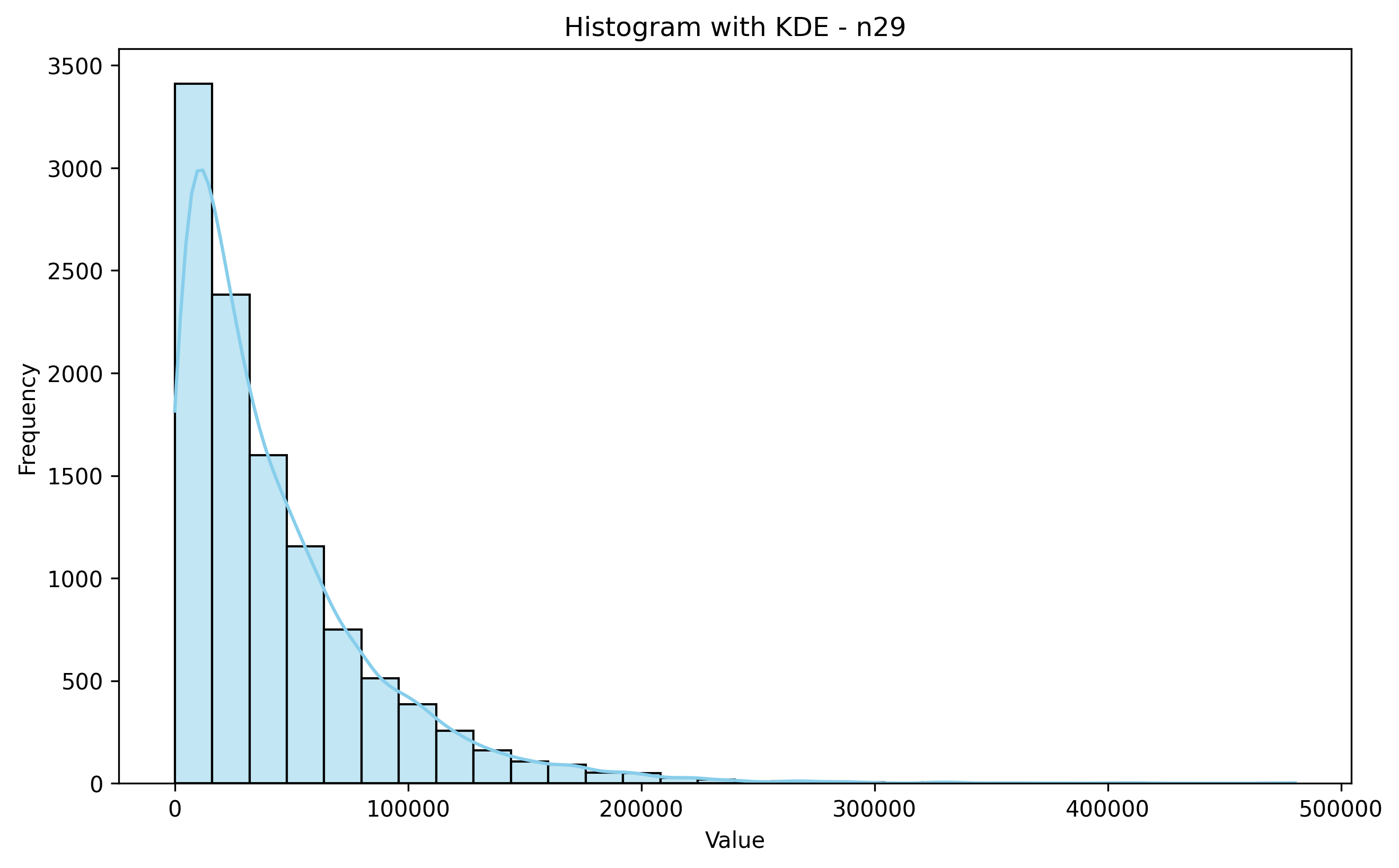}
    \caption{Las Vegas Attempt distribution for n = 29}
\end{subfigure}
\hfill
\begin{subfigure}[b]{0.3\textwidth}
    \includegraphics[width=\textwidth]{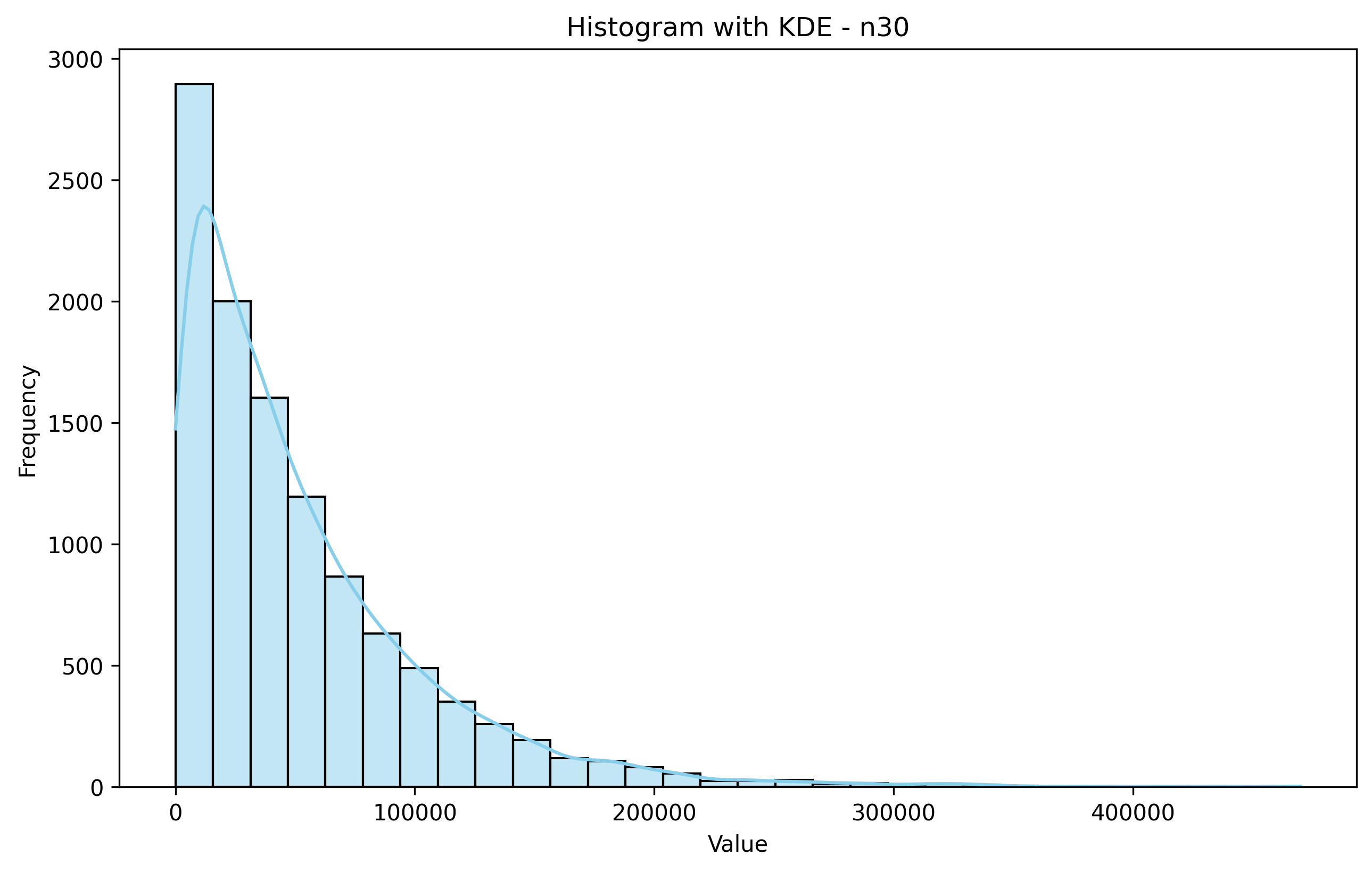}
    \caption{Las Vegas Attempt distribution for n = 30}
\end{subfigure}

\begin{subfigure}[b]{0.3\textwidth}
    \includegraphics[width=\textwidth]{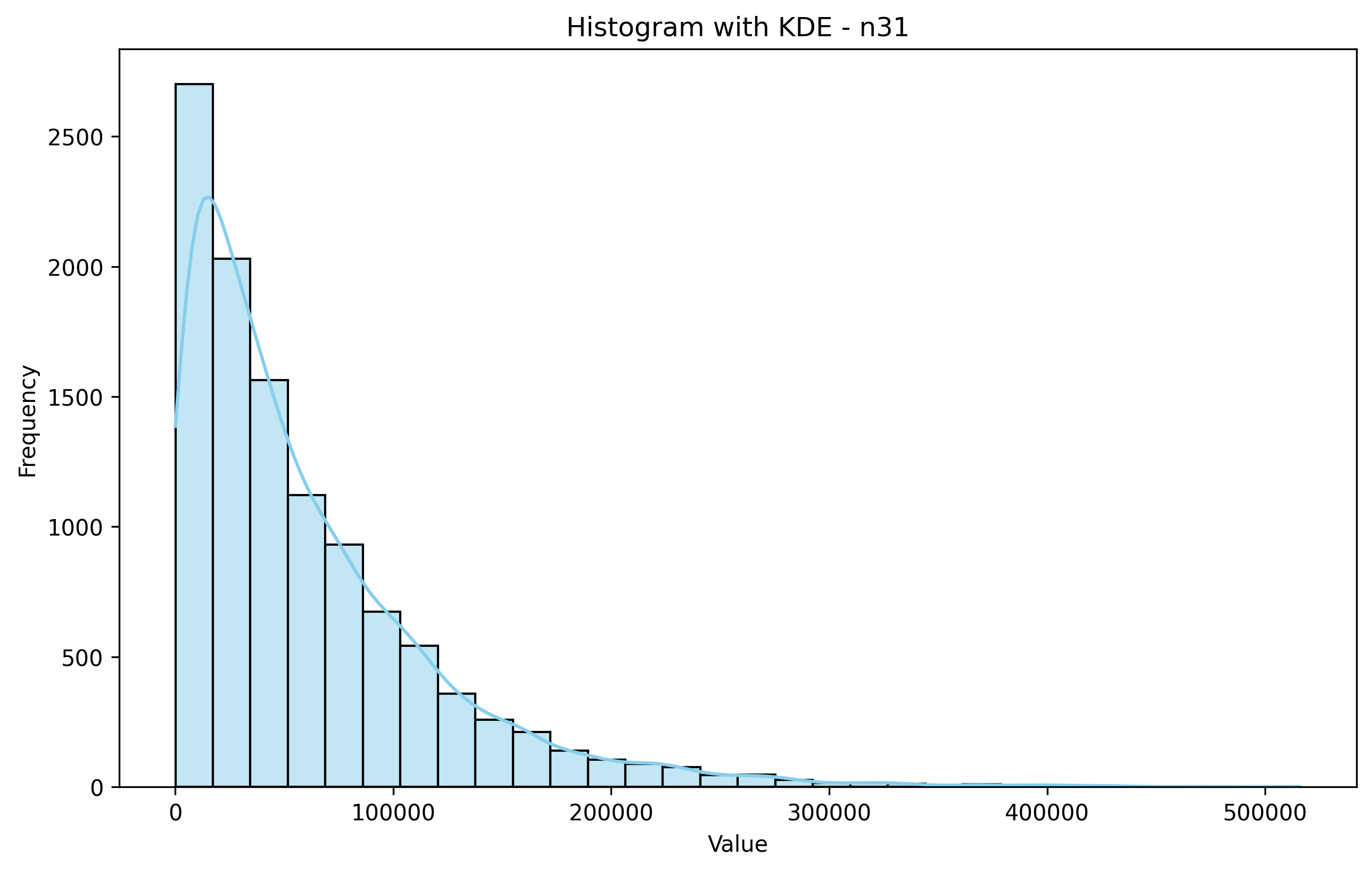}
    \caption{Las Vegas Attempt distribution for n = 31}
\end{subfigure}
\hfill
\begin{subfigure}[b]{0.3\textwidth}
    \includegraphics[width=\textwidth]{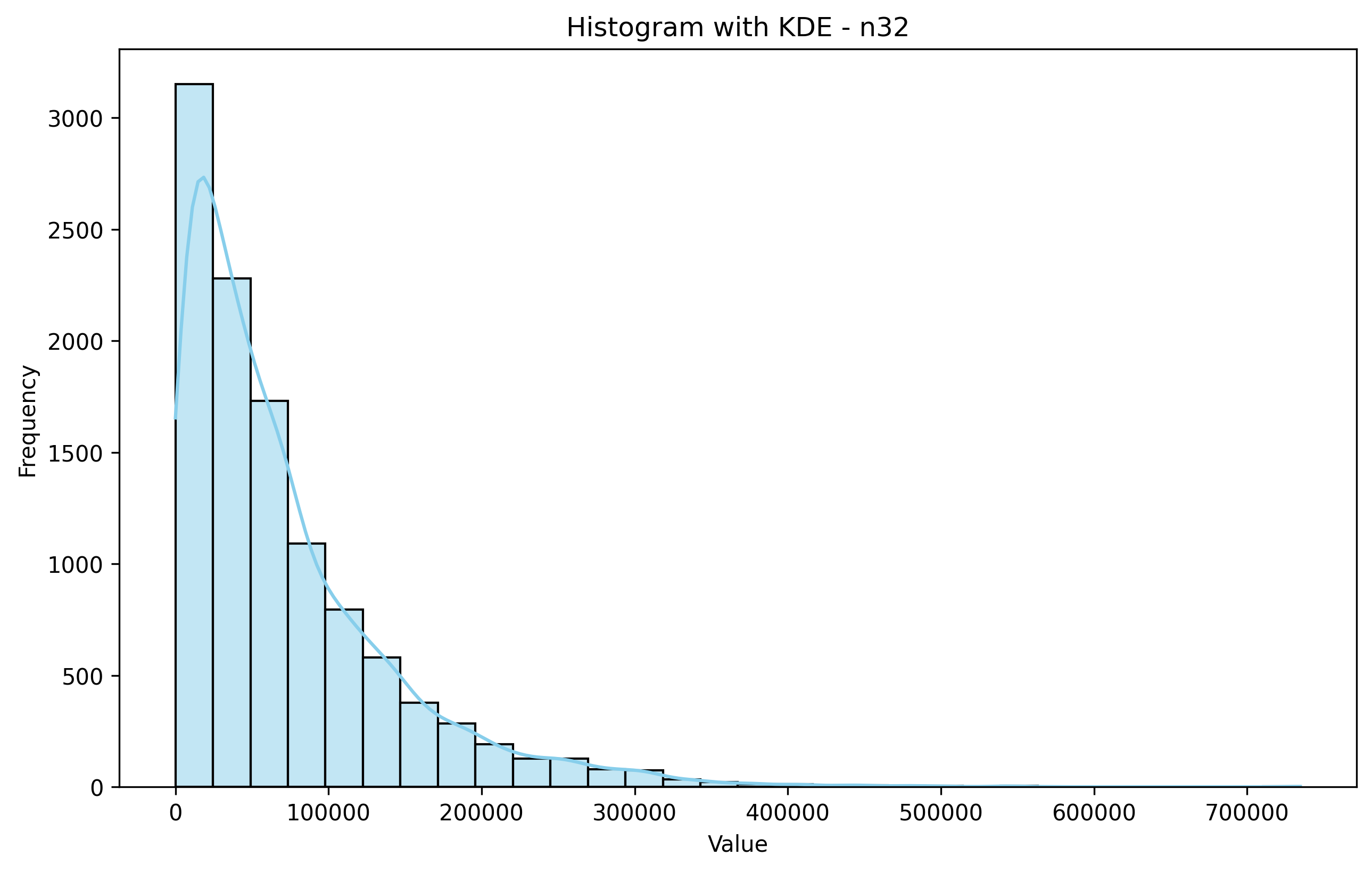}
    \caption{Las Vegas Attempt distribution for n = 32}
\end{subfigure}
\hfill
\begin{subfigure}[b]{0.3\textwidth}
    \includegraphics[width=\textwidth]{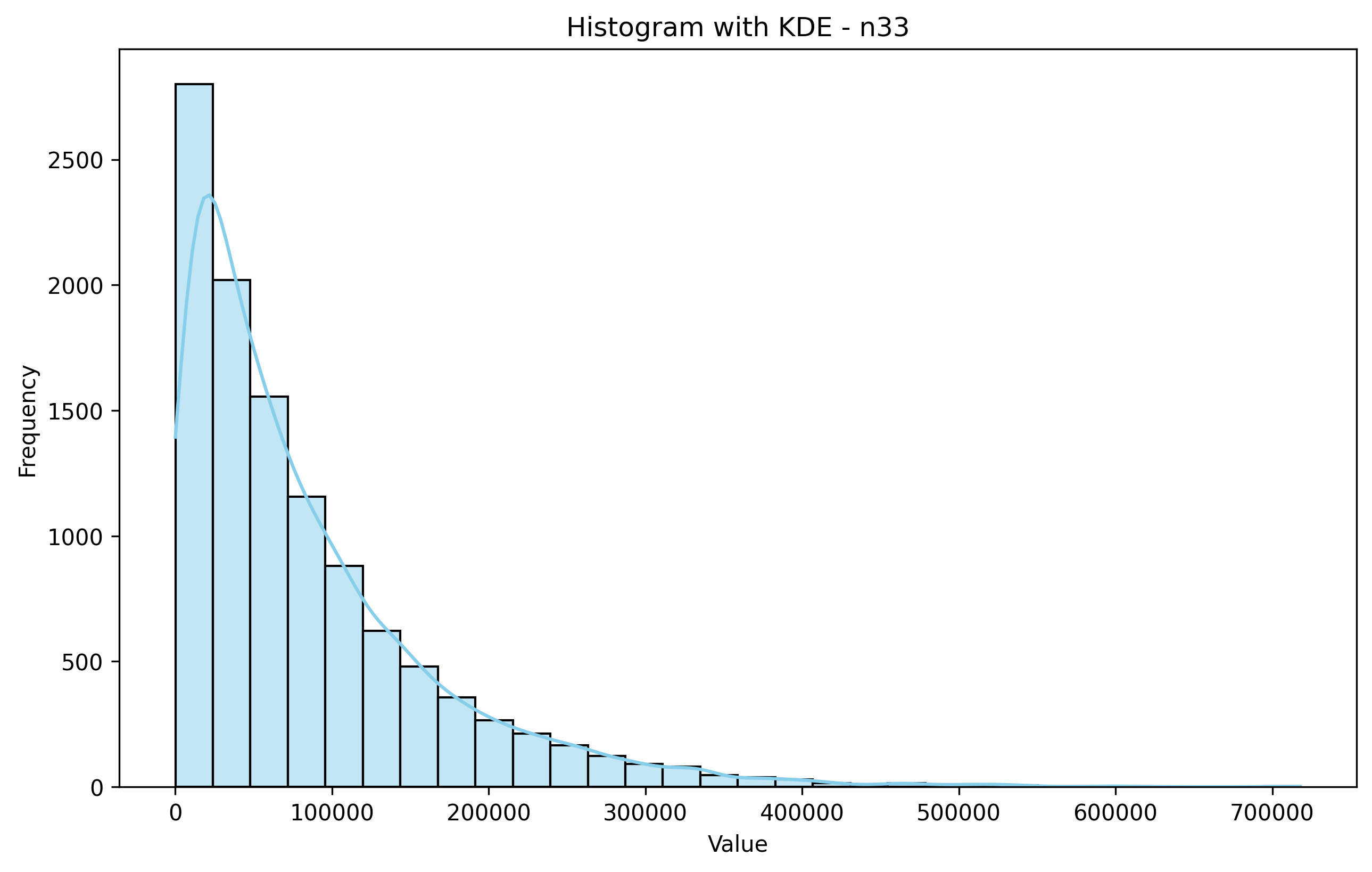}
    \caption{Las Vegas Attempt distribution for n = 33}
\end{subfigure}

\begin{subfigure}[b]{0.3\textwidth}
    \includegraphics[width=\textwidth]{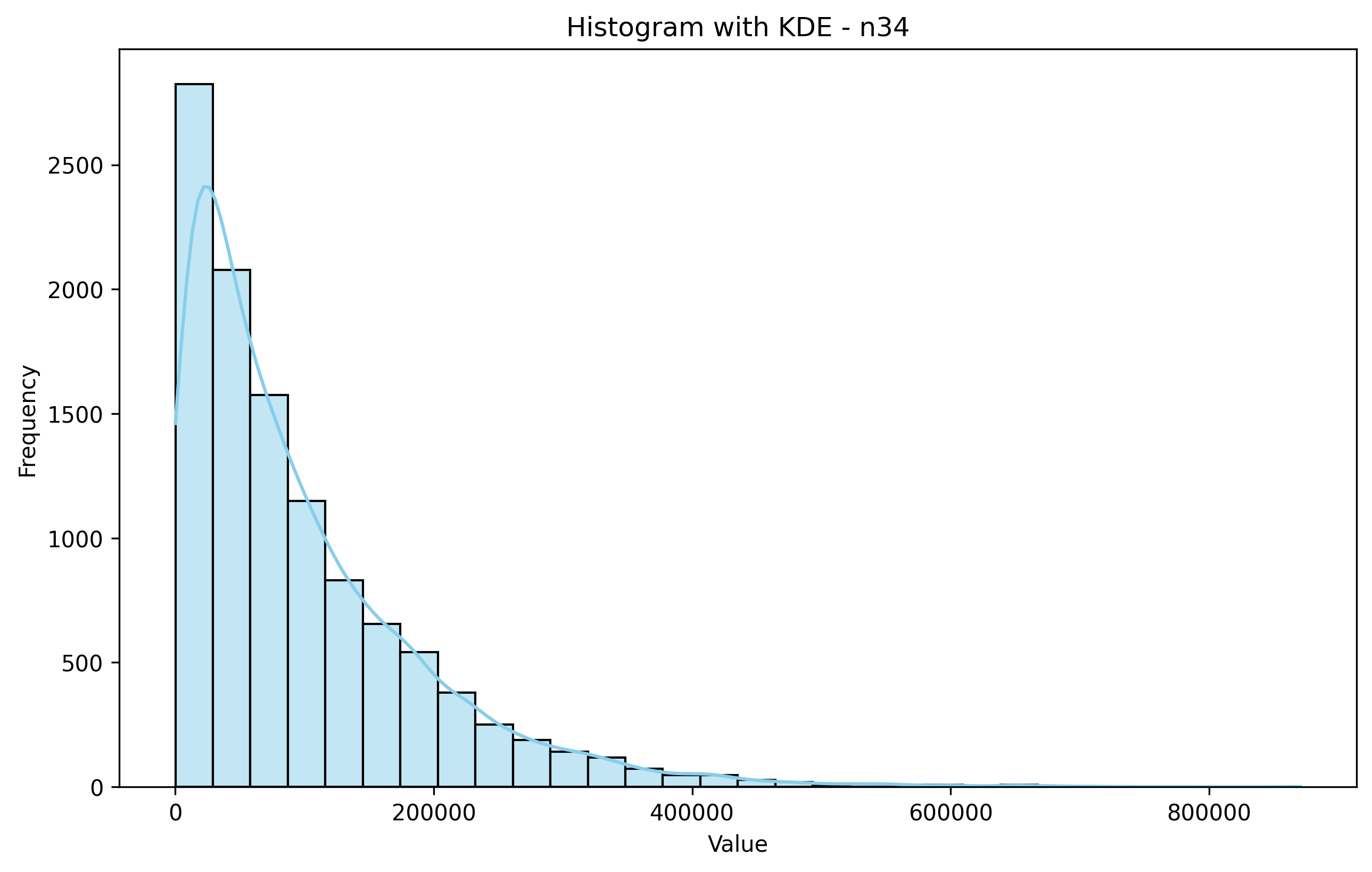}
    \caption{Las Vegas Attempt distribution for n = 34}
\end{subfigure}
\hfill
\begin{subfigure}[b]{0.3\textwidth}
    \includegraphics[width=\textwidth]{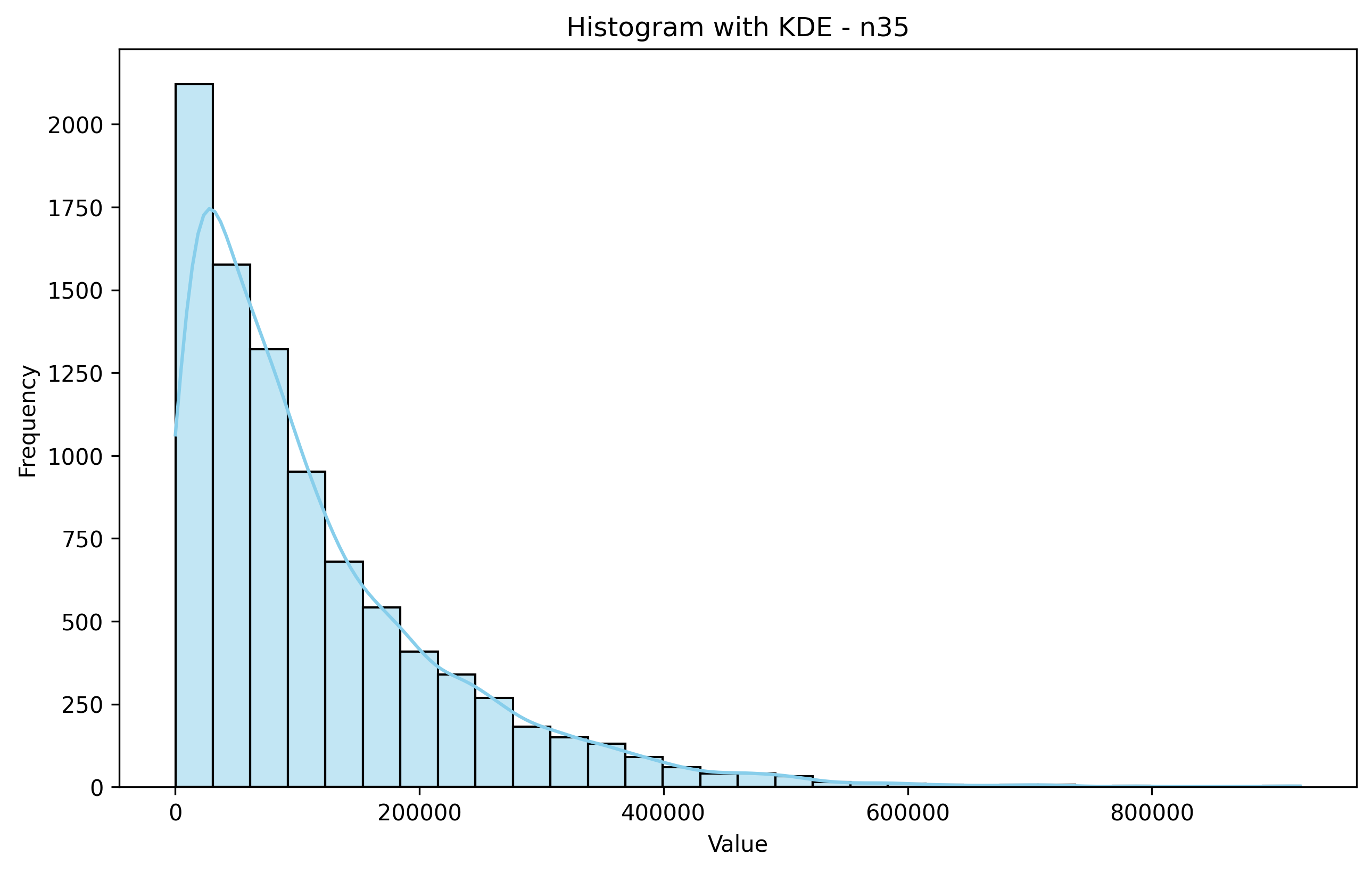}
    \caption{Las Vegas Attempt distribution for n = 35}
\end{subfigure}
\hfill
\begin{subfigure}[b]{0.3\textwidth}
    \includegraphics[width=\textwidth]{n36_histogram.png}
    \caption{Las Vegas Attempt distribution for n = 36}
\end{subfigure}

\begin{subfigure}[b]{0.3\textwidth}
    \includegraphics[width=\textwidth]{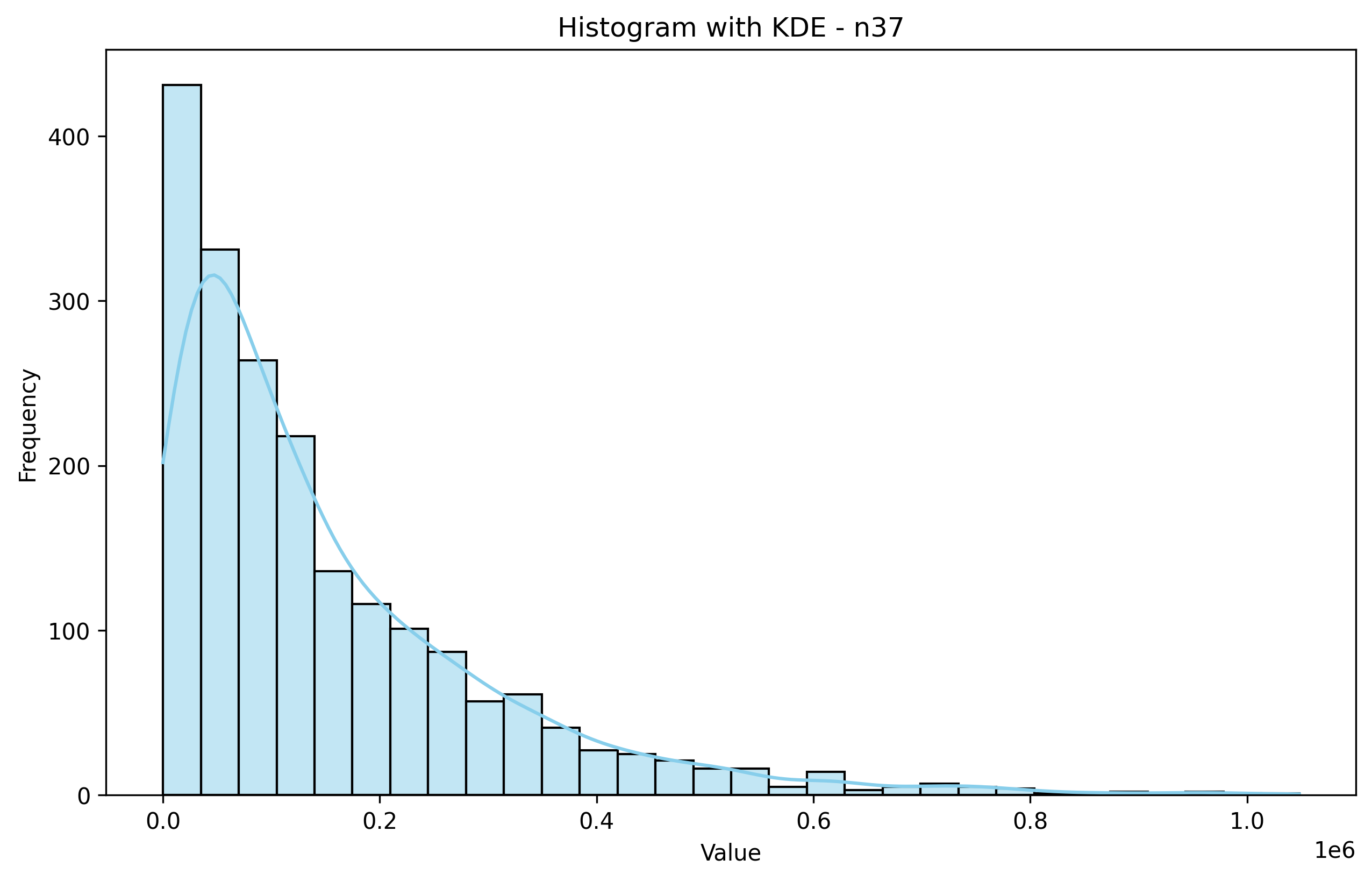}
    \caption{Las Vegas Attempt distribution for n = 37}
\end{subfigure}
\hfill
\begin{subfigure}[b]{0.3\textwidth}
    \includegraphics[width=\textwidth]{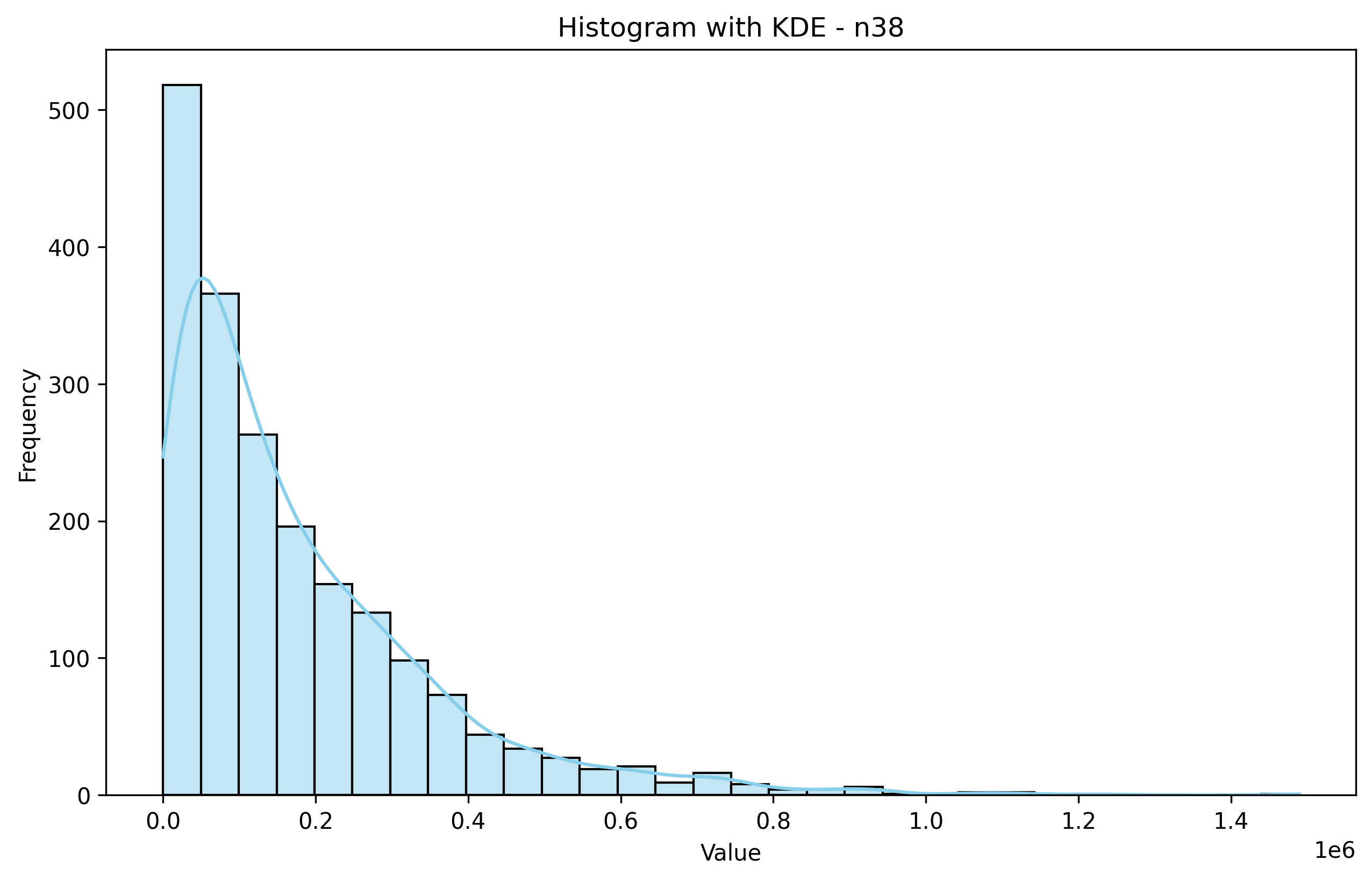}
    \caption{Las Vegas Attempt distribution for n = 38}
\end{subfigure}
\hfill
\begin{subfigure}[b]{0.3\textwidth}
    \includegraphics[width=\textwidth]{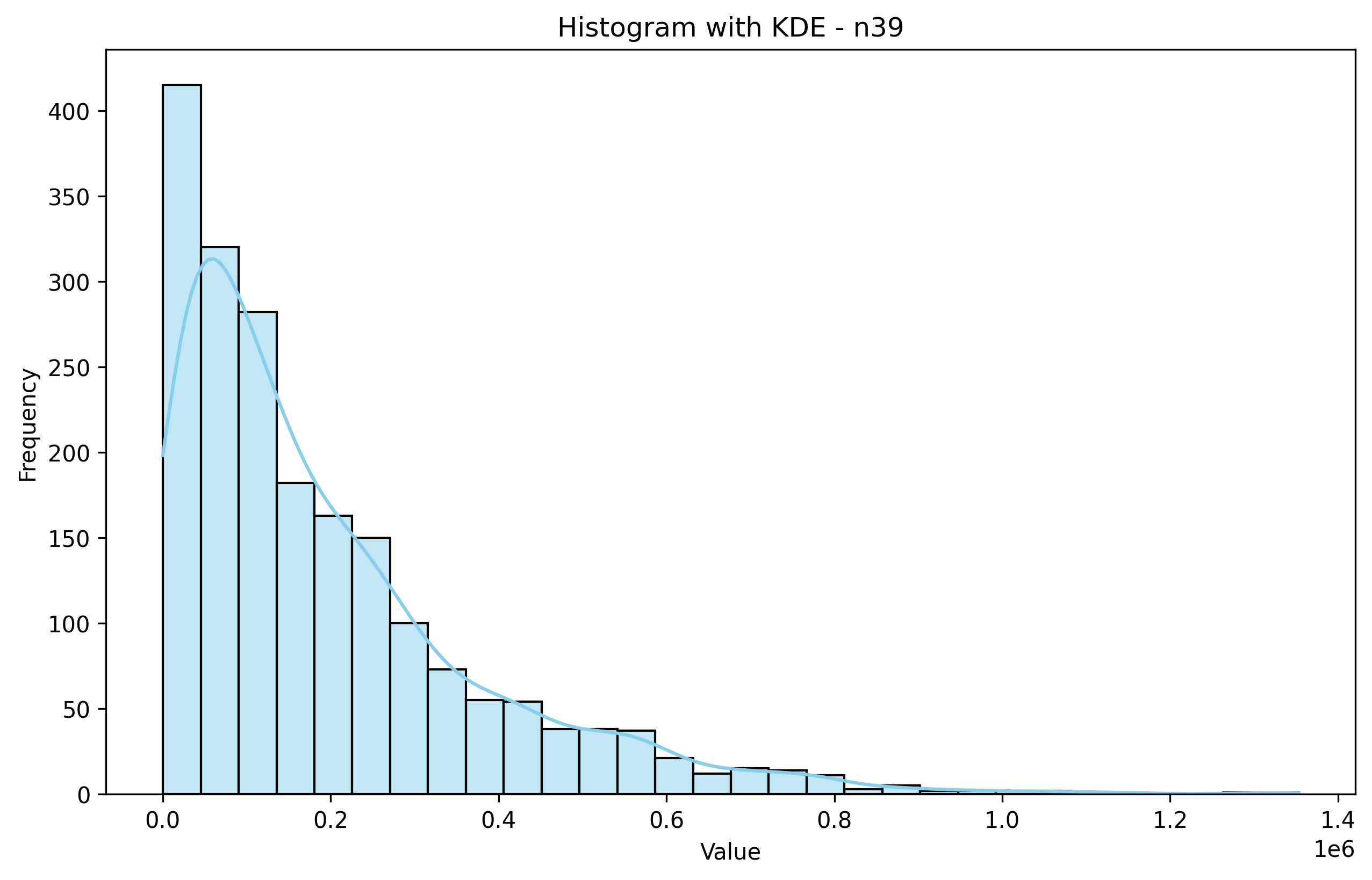}
    \caption{Las Vegas Attempt distribution for n = 39}
\end{subfigure}

\caption{Las Vegas Algorithm Attempt Distributions for Various Values of n}
\label{fig:las_vegas_distributions}
\end{figure}

One of the nqueen solution visulaized by Algorithm \ref{alg:las_vegas}

\begin{figure}[H]
  \centering
  \includegraphics[width=0.6\textwidth]{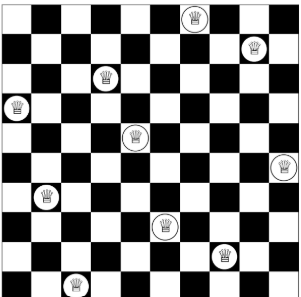}
  \caption{Nqueen Solution visualized}
  \label{fig:supp1}
\end{figure}

\subsection{Supplementary Data} \label{app:data}
Table~\ref{tab:appendix_data} provides detailed results comparing the probabilistic Las Vegas algorithm and the classical backtracking method for the $n$-Queens problem.
\begin{table}[H]
\centering
\caption{Summary of performance metrics for Las Vegas and Backtracking algorithms}
\label{tab:appendix_data}
\rotatebox{90}{
\begin{tabular}{ccccccccccc}
\toprule
 $n$ &  Mean &  Median &  Mode &  Skew &  Kurtosis &  Lower &  Upper & Distribution &  BackAttempts &  Speedup \\
\midrule
   4 &          16.585 &              13.0 &             4.0 &               2.077 &               6.289 &                   4.00 &                  46.00 &            weibullmin &                    26.0 &           1.568 \\
   5 &           9.792 &               9.0 &             5.0 &               1.939 &               5.124 &                   5.00 &                  23.00 &                  gamma &                    15.0 &           1.532 \\
   6 &         129.499 &              93.0 &             6.0 &               1.974 &               5.793 &                  11.00 &                 379.05 &            weibullmin &                   171.0 &           1.320 \\
   7 &          44.293 &              31.0 &             7.0 &               1.958 &               5.271 &                   7.00 &                 125.00 &              exponweib &                    42.0 &           0.948 \\
   8 &         103.268 &              74.0 &             8.0 &               2.013 &               6.044 &                   8.00 &                 298.00 &                  gamma &                   876.0 &           8.483 \\
   9 &         142.045 &             100.0 &             9.0 &               2.022 &               6.662 &                  15.00 &                 404.00 &                 pareto &                   333.0 &           2.344 \\
  10 &         400.540 &             280.0 &            10.0 &               1.942 &               5.577 &                  26.00 &                1157.00 &                   beta &                   975.0 &           2.434 \\
  11 &         675.372 &             470.0 &            11.0 &               2.005 &               5.759 &                  39.00 &                1991.00 &                 pareto &                   517.0 &           0.766 \\
  12 &         825.723 &             575.0 &            12.0 &               1.964 &               5.377 &                  51.00 &                2479.15 &                 pareto &                  3066.0 &           3.713 \\
  13 &        1083.545 &             761.0 &            13.0 &               1.931 &               5.217 &                  68.00 &                3200.15 &                 pareto &                  1365.0 &           1.260 \\
  14 &        1695.009 &            1180.0 &            14.0 &               2.004 &               5.755 &                 102.90 &                5018.40 &                 pareto &                 26495.0 &          15.631 \\
  15 &        2103.700 &            1480.0 &            15.0 &               2.169 &               8.094 &                 116.00 &                6266.85 &                 pareto &                 20280.0 &           9.640 \\
  16 &        2687.235 &            1843.0 &            16.0 &               2.015 &               5.668 &                 141.00 &                8068.70 &                 pareto &                160712.0 &          59.806 \\
  17 &        3520.288 &            2423.0 &            17.0 &               2.118 &               6.796 &                 180.00 &               10502.50 &                   beta &                 91222.0 &          25.913 \\
  18 &        4507.133 &            3185.5 &            18.0 &               1.814 &               4.342 &                 236.00 &               13528.60 &                 pareto &                743229.0 &         164.901 \\
  19 &        5659.341 &            3866.0 &            19.0 &               2.096 &               6.403 &                 282.90 &               17572.80 &                 pareto &                 48184.0 &           8.514 \\
  20 &        7286.821 &            5002.5 &            20.0 &               2.110 &               7.395 &                 382.95 &               21716.00 &                 pareto &               3992510.0 &         547.908 \\
  21 &        9040.502 &            6123.0 &            21.0 &               1.951 &               5.144 &                 503.00 &               27362.10 &                 pareto &                179592.0 &          19.865 \\
  22 &       11493.400 &            7848.0 &            22.0 &               2.011 &               5.911 &                 607.95 &               34561.30 &                 pareto &              38217905.0 &        3325.204 \\
  23 &       13774.240 &            9600.5 &            23.0 &               2.060 &               6.517 &                 712.85 &               41597.40 &                 pareto &                584591.0 &          42.441 \\
  24 &       16788.220 &           11659.5 &            24.0 &               1.946 &               5.387 &                 877.95 &               49734.80 &            weibullmin &               9878316.0 &         588.408 \\
  25 &       20931.828 &           14566.0 &            25.0 &               2.013 &               5.973 &                1086.95 &               62414.75 &                 pareto &               1216775.0 &          58.130 \\
  26 &       25420.719 &           17926.0 &            26.0 &               1.879 &               4.664 &                1377.00 &               77466.60 &            weibullmin &              10339849.0 &         406.749 \\
  27 &       30357.201 &           21483.0 &          3715.0 &               1.849 &               4.711 &                1618.00 &               89049.90 &                   beta &              12263400.0 &         403.970 \\
  28 &       35609.765 &           24825.5 &            28.0 &               1.940 &               5.529 &                1789.85 &              104913.20 &                 pareto &              84175966.0 &        2363.845 \\
  29 &       43147.836 &           29563.0 &            29.0 &               2.206 &               8.114 &                2290.90 &              128183.70 &              exponweib &              44434525.0 &        1029.820 \\
  30 &       52165.315 &           36444.0 &          5843.0 &               2.061 &               6.422 &                2802.60 &              153320.00 &                 pareto &            1692888135.0 &       32452.371 \\
  31 &       60285.698 &           42083.0 &         13214.0 &               1.937 &               5.264 &                3212.50 &              179305.00 &                   beta &             408773285.0 &        6780.601 \\
  32 &       71343.176 &           49718.5 &          8905.0 &               1.997 &               5.975 &                3741.95 &              216090.00 &                   beta &            2799725104.0 &       39243.068 \\
  33 &       83359.898 &           56972.0 &         17726.0 &               1.918 &               5.080 &                4617.00 &              252917.20 &              exponweib &             323601164.0 &        3881.976 \\
  34 &       96959.398 &           68097.0 &            34.0 &               1.851 &               4.881 &                4795.50 &              286823.20 &                  gamma &             707167767.0 &        7293.442 \\
  35 &      113575.668 &           78897.5 &            35.0 &               1.871 &               4.998 &                5882.30 &              340913.50 &            weibullmin &            3949267199.0 &       34772.124 \\
\bottomrule
\end{tabular}
}
\end{table}
\end{appendices}


\newpage


\section{References}
\begin{thebibliography}{20}
\bibitem{AlGburi2018} Al-Gburi, A. F. J., Naim, S., \& Boraik, A. N. (2018). Hybridization of bat and genetic algorithm to solve N-queens problem. \textit{Bulletin of Electrical Engineering and Informatics}, 7(4), 626--632. \href{https://doi.org/10.11591/eei.v7i4.1351}{https://doi.org/10.11591/eei.v7i4.1351}
\bibitem{Arteaga2022} Arteaga, A., Orozco-Rosas, U., Montiel, O., \& Castillo, O. (2022). Evaluation and comparison of brute-force search and constrained optimization algorithms to solve the N-queens problem. In O. Castillo \& P. Melin (Eds.), \textit{New perspectives on hybrid intelligent system design based on fuzzy logic, neural networks and metaheuristics} (pp. 121--140). Springer. \href{https://doi.org/10.1007/978-3-031-08266-5_9}{https://doi.org/10.1007/978-3-031-08266-5\_9}
\bibitem{Bell2009} Bell, J., \& Stevens, B. (2009). A survey of known results and research areas for n-queens. \textit{Discrete Mathematics}, 309(1), 1--31. \href{https://doi.org/10.1016/j.disc.2007.12.043}{https://doi.org/10.1016/j.disc.2007.12.043}
\bibitem{Bernhardsson1991} Bernhardsson, B. (1991). Explicit solutions to the n-queens problem for all n. \textit{ACM SIGART Bulletin}, 2(2), 7. \href{https://doi.org/10.1145/122319.122322}{https://doi.org/10.1145/122319.122322}
\bibitem{Erbas1992} Erbaş, C., Sarkeshik, S., \& Tanik, M. M. (1992). Different perspectives of the N-queens problem. \textit{Proceedings of the 1992 ACM Annual Conference on Communications}, 99--108. \href{https://doi.org/10.1145/131214.131227}{https://doi.org/10.1145/131214.131227}
\bibitem{Gent2017} Gent, I. P., Jefferson, C., \& Nightingale, P. (2017). Complexity of n-queens completion. \textit{Journal of Artificial Intelligence Research}, 59, 815--848. \href{https://doi.org/10.1613/jair.5512}{https://doi.org/10.1613/jair.5512}
\bibitem{Gomes2008} Gomes, C. P., Kautz, H., Sabharwal, A., \& Selman, B. (2008). Satisfiability solvers. In F. van Harmelen, V. Lifschitz, \& B. Porter (Eds.), \textit{Handbook of knowledge representation} (pp. 89--134). Elsevier. \href{https://doi.org/10.1016/S1574-6526(07)03002-7}{https://doi.org/10.1016/S1574-6526(07)03002-7}
\bibitem{Haralick1980} Haralick, R. M., \& Elliott, G. L. (1980). Increasing tree search efficiency for constraint satisfaction problems. \textit{Artificial Intelligence}, 14(3), 263--313. \href{https://doi.org/10.1016/0004-3702(80)90051-X}{https://doi.org/10.1016/0004-3702(80)90051-X}
\bibitem{Hoffman1969} Hoffman, E. J., Loessi, J. C., \& Moore, R. C. (1969). Constructions for the solution of the m-queens problem. \textit{Mathematics Magazine}, 42(2), 66--72. \href{https://doi.org/10.1080/0025570X.1969.11975924}{https://doi.org/10.1080/0025570X.1969.11975924}
\bibitem{Levitin2012} Levitin, A. (2012). \textit{Introduction to the design \& analysis of algorithms} (3rd ed.). Pearson. \href{https://www.pearson.com/en-us/subject-catalog/p/introduction-to-the-design-and-analysis-of-algorithms/P200000003500/9780137545186}{https://www.pearson.com/en-us/subject-catalog/p/introduction-to-the-design-and-analysis-of-algorithms/P200000003500/9780137545186}
\bibitem{Luria2021} Luria, Z., \& Simkin, M. (2021). A lower bound for the n-queens problem. \textit{arXiv preprint arXiv:2105.11431}. \href{https://arxiv.org/abs/2105.11431}{https://arxiv.org/abs/2105.11431}
\bibitem{McConnell2001} McConnell, J. J. (2001). \textit{Analysis of algorithms: An active learning approach}. Jones and Bartlett Publishers. \href{https://dl.acm.org/doi/10.5555/517060}{https://dl.acm.org/doi/10.5555/517060}
\bibitem{Minton1992} Minton, S., Johnston, M. D., Philips, A. B., \& Laird, P. (1992). Minimizing conflicts: A heuristic repair method for constraint satisfaction and scheduling problems. \textit{Artificial Intelligence}, 58(1-3), 161--205. \href{https://doi.org/10.1016/0004-3702(92)90007-K}{https://doi.org/10.1016/0004-3702(92)90007-K}
\bibitem{Polson2024} Polson, N. G., \& Sokolov, V. (2024). Counting N-queens: Monte Carlo methods in polynomial time. \textit{arXiv preprint arXiv:2407.08830}. \href{https://arxiv.org/abs/2407.08830}{https://arxiv.org/abs/2407.08830}
\bibitem{Rolfe2006} Rolfe, T. J. (2006). Las Vegas does n-queens. \textit{ACM SIGCSE Bulletin}, 38(2), 37--38. \href{https://doi.org/10.1145/1138403.1138429}{https://doi.org/10.1145/1138403.1138429}
\bibitem{Russell2020} Russell, S. J., \& Norvig, P. (2020). \textit{Artificial intelligence: A modern approach} (4th ed.). Pearson. \href{https://aima.cs.berkeley.edu/}{https://aima.cs.berkeley.edu/}
\bibitem{Sosic1990} Sosic, R., \& Gu, J. (1990). A polynomial time algorithm for the n-queens problem. \textit{ACM SIGART Bulletin}, 1(3), 7--11. \href{https://doi.org/10.1145/101340.101343}{https://doi.org/10.1145/101340.101343}
\bibitem{Sosic1991} Sosic, R., \& Gu, J. (1991). Fast search algorithms for the n-queens problem. \textit{IEEE Transactions on Systems, Man, and Cybernetics}, 21(6), 1572--1576. \href{https://doi.org/10.1109/21.135698}{https://doi.org/10.1109/21.135698}
\bibitem{Sosic1994} Sosic, R., \& Gu, J. (1994). Efficient local search with conflict minimization: A case study of the n-queens problem. \textit{IEEE Transactions on Knowledge and Data Engineering}, 6(5), 661--668. \href{https://doi.org/10.1109/69.317698}{https://doi.org/10.1109/69.317698}
\bibitem{vanBeek2006} van Beek, P. (2006). Backtracking search algorithms. In F. Rossi, P. van Beek, \& T. Walsh (Eds.), \textit{Handbook of constraint programming} (pp. 85--134). Elsevier. \href{https://doi.org/10.1016/S1574-6526(06)80008-8}{https://doi.org/10.1016/S1574-6526(06)80008-8}
\bibitem{Wu2006} Wu, H. (2006). \textit{Randomization and restart strategies} [Master's thesis]. University of Waterloo. \href{https://www.collectionscanada.gc.ca/obj/s4/f2/dsk3/OWTU/TC-OWTU-992.pdf}{https://www.collectionscanada.gc.ca/obj/s4/f2/dsk3/OWTU/TC-OWTU-992.pdf}
\end{thebibliography}
\end{document}